\pdfoutput=1 
\documentclass[runningheads]{llncs}
\usepackage{graphicx}

\usepackage{tikz}
\usepackage{comment}
\usepackage{amsmath,amssymb} 
\usepackage{color}

\usepackage[accsupp]{axessibility}  

\usepackage[width=122mm,left=20mm,paperwidth=162mm,height=193mm,top=20mm,paperheight=233mm]{geometry}

\usepackage{graphicx}
\usepackage{amsmath}
\usepackage{amssymb}
\usepackage{booktabs}

\usepackage{microtype}
\usepackage{graphicx}

\usepackage{times}
\usepackage{epsfig}
\usepackage{comment}
\usepackage{dsfont}
\usepackage{xspace}
\usepackage{multirow}
\usepackage{caption}

\usepackage{adjustbox}
\usepackage{framed}
\usepackage{algorithm,algpseudocode}
\usepackage{bibunits}
\usepackage{tabularx}
\usepackage{subcaption}
\usepackage{framed}
\usepackage{comment}
\usepackage{wrapfig,booktabs}
\usepackage{float}

\usepackage[pagebackref,breaklinks,colorlinks]{hyperref}
\usepackage{soul}
\usepackage{xcolor}

\usepackage{enumitem}
\setlist{itemsep=0.2em,topsep=0.3em,parsep=0pt,partopsep=0pt,leftmargin=1.2em}

\usepackage[capitalize]{cleveref}
\crefname{section}{Sec.}{Secs.}
\Crefname{section}{Section}{Sections}
\Crefname{table}{Table}{Tables}
\crefname{table}{Tab.}{Tabs.}

\usepackage[normalem]{ulem}
\usepackage{soul}
\usepackage{xstring}

\sethlcolor{magenta}
\makeatletter
\def\SOUL@hlpreamble{%
    \setul{\dp\strutbox}{\dimexpr\ht\strutbox+\dp\strutbox\relax}%
    \let\SOUL@stcolor\SOUL@hlcolor
    \SOUL@stpreamble
}
\makeatother

\newcommand{\addref}[1]{\textcolor{magenta}{ [REF] }}

\DeclareMathOperator*{\aggre}{\text{AGGRE}\hspace{0.3em}}
\DeclareMathOperator*{\concat}{\|}

\definecolor{mygray}{gray}{0.95}

\newtheorem{remark1}{Remark}%

\begin{document}
\pagestyle{headings}
\mainmatter
\def\ECCVSubNumber{6067}  

\title{3D Equivariant Graph Implicit Functions
} 

\begin{comment}
\titlerunning{ECCV-22 submission ID \ECCVSubNumber} 
\authorrunning{ECCV-22 submission ID \ECCVSubNumber} 
\author{Anonymous ECCV submission}
\institute{Paper ID \ECCVSubNumber}
\end{comment}

\titlerunning{3D Equivariant Graph Implicit Functions}
%
\author{Yunlu Chen\inst{1} \and
Basura Fernando\inst{2} \and
Hakan Bilen\inst{3} \and \\
Matthias Nie{\ss}ner\inst{4} \and
Efstratios Gavves\inst{1}
}
\authorrunning{Y. Chen et al.}
%
\institute{University of Amsterdam 
\and CFAR, IHPC, A*STAR 
\and
University of Edinburgh 
\and Technical University of Munich 
\\ \vspace{0.5em} \email{\{y.chen3, e.gavves\}@uva.nl, fernandopbc@ihpc.a-star.edu.sg, hbilen@ed.ac.uk, niessner@tum.de}
}
\maketitle

\begin{abstract}

In recent years, neural implicit representations have made remarkable progress in modeling of 3D shapes with arbitrary topology.
In this work, we address two key limitations of such representations, in failing to capture local 3D geometric fine details, and to learn from and generalize to shapes with unseen 3D transformations.
To this end, we introduce a novel family of graph implicit functions with equivariant layers that facilitates modeling fine local details and guaranteed robustness to various groups of geometric transformations, through local $k$-NN graph embeddings with sparse point set observations at multiple resolutions.
Our method improves over the existing rotation-equivariant implicit function from 0.69 to 0.89 (IoU) on the ShapeNet reconstruction task.
We also show that our equivariant implicit function can be extended to other types of similarity transformations and generalizes to unseen translations and scaling.
\end{abstract}

\begin{figure}[t]
\centering 
\vspace{-0.6em}
\includegraphics[trim={6cm 0 0 0},clip,width=0.72\linewidth]{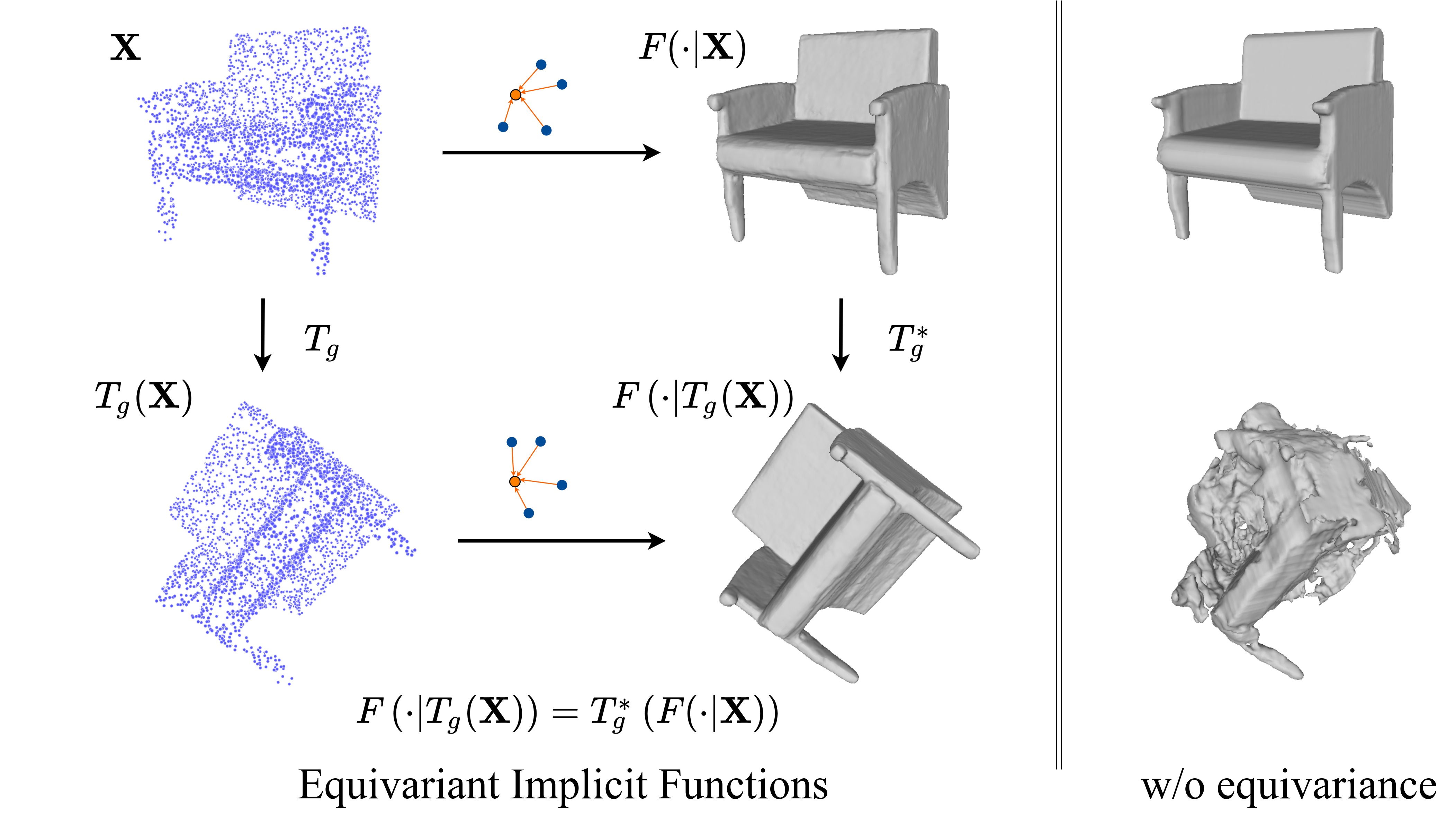}
\caption{\textbf{Our equivariant graph implicit function} infers the implicit field $F(\cdot| \mathbf{X})$ for a 3D shape, given a sparse point cloud observation $\mathbf{X}$. When a transformation $T_g$ (rotation, translation, or/and scaling) is applied to the observation $\mathbf{X}$, the resulting implicit field  $F(\cdot|T_g(\mathbf{X}))$ is {\it guaranteed} to be the same as applying a corresponding transformation $T_g^\ast$ to the inferred implicit field from the untransformed input (\textit{middle}). The property of equivariance enables generalization to unseen transformations, under which existing models such as ConvONet~\cite{peng2020convolutional} often struggle ({\it right}).
}
\vspace{-0.3em}
\label{fig:splash} 
\end{figure}

\section{Introduction}

Neural implicit representations are effective at encoding 3D shapes of arbitrary topology~\cite{park2019deepsdf,mescheder2019occnet,chen2019imnet}. Their key idea is to represent a shape by a given latent code in the learned manifold and for each point in space, the neural implicit function checks whether a given coordinate location is occupied within the shape or not. In contrast to traditional discrete 3D representations such as triangle meshes or point clouds, this new paradigm of implicit neural representations has gained significant popularity due to the advantages such as being continuous, grid-free, and the ability to handle various topologies.

Despite their success, latent-code-conditioned implicit representations have two key limitations. First, the latent code of the shape captures coarse high-level shape details (i.e., the global structure) without any explicit local spatial information, hence it is not possible to learn correlations between the latent code and local 3D structural details of the shape. As a result, the surface reconstruction from latent-code-conditioned implicit functions tends to be over-smoothed and they are not good at capturing local surface detail~\cite{peng2020convolutional,chibane2020ifnet,jiang2020lig,genova2019deep,erler2020points2surf}.
Second, implicit representations are sensitive to various  geometric transformations, in particular to rotations~\cite{davies2020effectiveness}. The performance of the implicit representations heavily relies on the assumption that shape instances in the same category are required to be in the same canonical orientation such that shape structures of planes and edges are in line with the coordinate axes.
While data augmentation loosely addresses this second issue to some degree, a principled approach is to enable the representations to be inherently aware of common geometric operations such as rotations, translations, and scaling, which are found commonly in real-world 3D objects.

To address the first challenge of modeling local spatial information, recent methods~\cite{peng2020convolutional,chibane2020ifnet} first discretize 3D space into local 2D or 3D grids and then store implicit codes locally in the respective grid cells. 
However, these methods are still sensitive to transformations as the grid structure is constructed in line with the chosen coordinate axes. This results in deteriorated performance under transformations as shown in Fig.~\ref{fig:splash}. 
In addition, the grid discretization often has to trade fine details of the shape, hence the quality of shape reconstruction, for better computational efficiency through a low resolution grid.
Deng \textit{et al.}~\cite{deng2021vectorneurons} propose VN-ONet to tackle the second challenge of pose-sensitivity with a novel vector neuron formulation, which enables the network architecture to have a rotation equivariant representation.
Nevertheless, similar to implicit representations, the VN-ONet encodes each shape with a global representation and hence fails to capture local details. As grid discretization is not robust to transformations, this solution is not compatible with the grid-based local implicit methods. 
Thus, integrating VN-ONet with grid-approach is not a feasible solution.

In this work, our goal is to simultaneously address both challenges of encoding local details in latent representations and dealing with the sensitivity to geometric transformations such as rotations, translations, and scaling.
To this end, we propose a novel equivariant graph-based local implicit function that unifies both of these properties.
{In particular, we use graph convolutions to capture local 3D information in a non-Euclidean manner, with a multi-scale sampling design in the architecture to aggregate global and local context at different sampling levels. We further integrate equivariant layers to facilitate generalization to unseen geometric transformations.}
Unlike the grid-based methods~\cite{peng2020convolutional,chibane2020ifnet} that requires discretization of 3D space into local grids, our graph-based implicit function uses point features from the input point cloud observation directly without interpolation from grid features. 
Our graph mechanism allows the model to attend detailed information from fine areas of the shape surface points, while the regularly-spanned grid frame may place computations to less important areas.
In addition, our graph structure is not biased towards the canonical axis directions of the given Cartesian coordinate frame, hence less sensitive than the grid-local representations~\cite{peng2020convolutional,chibane2020ifnet}.
Therefore, our graph representation is maximally capable of realizing an equivariant architecture for 3D shape representation.
In summary, our contributions are as follows:
\begin{itemize}
    \item {We propose a novel graph-based implicit representation network that enables effective encoding of local 3D information in a multi-scale sampling architecture, and thus modeling of high-fidelity local 3D geometric detail. Our model features a non-Euclidean graph representation that naturally adapts with geometric transformations.} 
    
    \item We incorporate equivariant graph layers in order to facilitate inherent robustness against geometric transformations. Together with the graph embedding, our equivariant implicit model significantly improves the reconstruction quality from the existing rotation equivariant implicit method~\cite{deng2021vectorneurons}.
    
    \item  We extend our implicit method to achieve a stronger equivariant model that handles more types of similarity transformations simultaneously with guaranteed perfect generalization, including rotation, translation and scaling.  
\end{itemize}

\section{Related Work}

\noindent\textbf{Implicit 3D representations.} Neural implicits have been shown to be highly effective for encoding continuous 3D signals of varying topology~\cite{park2019deepsdf,mescheder2019occnet,chen2019imnet}. 
Its variants have been used in order to reconstruct shapes from a single image~\cite{saito2019pifu,xu2019disn,xu2020ladybird}, or use weaker supervision for raw point clouds~\cite{atzmon2019controlling,atzmon2020sal,atzmon2020sal++} and 2D views~\cite{liu2019haoli,niemeyer2020differentiable,jiang2020sdfdiff}.

 \vspace{0.3em}
\noindent\textbf{Local latent implicit embeddings.}
ConvONet~\cite{peng2020convolutional} and IF-Net~\cite{chibane2020ifnet} concurrently propose to learn multi-scale local grid features with convolution layers to improve upon global latent implicit representations. %
Other variants of grid methods~\cite{jiang2020lig,chabra2020deeplocalshapes} takes no global cues, hence restricted by requiring additional priors during inference such as normals~\cite{jiang2020lig} or the partial implicit field~\cite{chabra2020deeplocalshapes}. 
While grid approaches are sensitive to rotations, non-grid local implicit embeddings are not well explored. 
E.g. \cite{erler2020points2surf} constructs local patches from the shape and apply one PointNet~\cite{qi2017pointnet} on each of them, while 
the boosted computation cost by two orders of magnitude from using numerous PointNets restricts applications on large-scale datasets e.g. ShapeNet~\cite{chang2015shapenet}. In contrast to these methods, we use a hierarchical graph embedding that effectively encodes multi-scale context to address the pose-sensitivity of grid approaches.

 \vspace{0.3em}
\noindent\textbf{Pose-sensitivity in implicit functions.} As generalization becomes a concern for latent-coded implicit functions~\cite{sitzmann2020metasdf,bautista2021generalization,chen2021neural}, Davies {\it  et al.}~\cite{davies2020effectiveness} first point out that the implicit 3D representations are biased towards canonical orientations. %
Deng {\it et al.}~\cite{deng2021vectorneurons} introduce a rotation equivariant implicit network VN-ONet that generalises to random unseen rotations, %
but yet with the restrictions from the global latent. 
As grid embeddings are sensitive to rotation, seeking a compatible local latent embedding is a non-trivial problem.

 \vspace{0.3em}
\noindent\textbf{Rotation equivariance with 3D vector features.}
Equivariance has drawn attention in deep learning models with inductive priors of physical symmetries, %
e.g., the success of ConvNets are attributed to translation equivariance. Advanced techniques are developed for equivariance to rotation~\cite{cohen2016steerable,worrall2017harmonic,thomas2018tfn}, scale~\cite{sosnovik2019scale,zhu2019scale} and permutation~\cite{zaheer2017deepsets,qi2017pointnet}. %
Recently, a new paradigm for rotation equivariance uses 3D vectors as neural features~\cite{deng2021vectorneurons,shen2020quaternion} with improved effectiveness and efficiency upon methods based on spherical harmonics~\cite{worrall2017harmonic,thomas2018tfn,weiler2018learning,weiler20183dsteerable,fuchs2020se}. 
Shen {\it et al.}~\cite{shen2020quaternion} first introduced pure quaternion features that are equivalent to 3D vectors. 
Deng {\it et al.}~\cite{deng2021vectorneurons} proposed a similar design with improved nonlinear layers.
Satorras {\it et al.}~\cite{satorras2021n} proposed to aggregate vector inputs in graph message passing, but without vector nonlinearities involved. 
Leveraging on existing work~\cite{shen2020quaternion,deng2021vectorneurons,satorras2021n}, we introduce hybrid vector and scalar neural features for better performance and efficiency. We also adapt the paradigm for scale equivariance, for the first time in literature.

\section{A Definition of Equivariance for Implicit Representations}
\label{sec:equiv}

While equivariance to common geometric transformations is widely studied for explicit representations of 2D and 3D data~\cite{cohen2016groupequivariant,worrall2017harmonic,thomas2018tfn}, the property for implicit representations that encode signals in a function space is more challenging since continuous queries are involved, yet an important problem for 3D reconstruction.
We discuss standard 3D implicit functions and then define equivariance for the representation.

\vspace{0.3em}
\noindent\textbf{3D implicit representations.}
We build our model on neural 3D occupancy field functions~\cite{mescheder2019occnet}, widely used as a shared implicit representation for a collection of 3D shapes.
Given an observation $\mathbf{X}\in \mathcal{X}$ of a 3D shape, 
the conditional implicit representation of the shape, $F(\cdot| \mathbf{X} ): \mathbb{R}^3 \to [0,1]$, is a 3D scalar field that maps the 3D Euclidean domain to occupancy probabilities, indicating whether there is a surface point at the coordinate. %
In this work, we consider $\mathbf{X}$ as a sparse 3D point cloud, such that the information from the observation is indifferent under an arbitrary global transformation, as required for equivariance.
For each 3D query coordinate $\vec{p} \in \mathbb{R}^3$, the conditioned implicit representation  is in the form of 
\begin{align}
    F(\vec{p}| \mathbf{X} ) = \Psi(\vec{p}, \vec{z}) = \Psi(\vec{p}, \Phi(\mathbf{X})),  %
    \label{eq:global_latent}
\end{align}
where $\Phi(\mathbf{X}) = \vec{z} \in \mathbb{R}^{C_{\vec{z}}}$ is the latent code in the form of a $C_{\vec{z}}$-dimensional vector from the observation $\mathbf{X}$, and $\Phi$ is the {\it latent feature extractor} that encodes the observation data $\mathbf{X}$.  
$\Psi$ is the implicit decoder %
implemented using a multi-layered perceptron with ReLU activations.   %
Occupancy probabilities are obtained by the final sigmoid activation function of the implicit decoder. 
Following \cite{mescheder2019occnet,peng2020convolutional}, the model is trained with binary cross-entropy loss supervised by  ground truth occupancy. 
The underlying shape surface 
is the 2-manifold $\{ \vec{p'} |F(\vec{p'}| \mathbf{X} ) = \tau \}$, %
where $\tau \in (0,1)$ is the surface decision boundary.

\vspace{0.3em}
\noindent\textbf{Preliminary of equivariance.} Consider a set of transformations $T_g: \mathcal{X} \to \mathcal{X}$ on a vector space $\mathcal{X}$ for $g\in G$, where $G$ is an abstract group. Formally, $T_g = T(g)$ where $T$ is a representation of group $G$, such that $\forall g, g' \in G, T(gg')=T(g)T(g')$. 
In the case that $G$ is the 3D rotation group $\text{SO}(3)$, $T_g$ instantiates a 3D rotation matrix for a rotation denoted by $g$. We say a function $\Xi: \mathcal{X} \to \mathcal{Y}$ is equivariant with regard to group $G$ if there exists $T_g^\ast: \mathcal{Y} \to \mathcal{Y}$ such that for all $g \in G$: $ T^\ast_g \circ \Xi = \Xi \circ T_g$. We refer to \cite{cohen2016groupequivariant,thomas2018tfn} for more detailed background theory. 
Next, we define equivariance for implicit representations as follows: 

\vspace{-0.3em}
\begin{definition}[Equivariant 3D implicit functions]
Given a group $G$ and the 3D transformations $T_g$ with $g\in G$, the conditioned implicit function $ F(\cdot| \mathbf{X})$ is equivariant with regard to $G$, if%
\begin{align}
\hspace{-0.8em}
 F\big(\cdot|T_g(\mathbf{X})\big) = T_g^\ast\big(F(\cdot|\mathbf{X})\big), %
 \quad \text{for all } g \in G, \mathbf{X} \in \mathcal{X}.
   \label{eq:equiv}
\end{align}
where the transformation $T_g^\ast$ applied on the implicit function associated to $T_g$ is applying the inverse coordinate transform on query coordinates $T_g^\ast(F(\cdot|\mathbf{X})) \equiv F(T_g^{-1}(\cdot)|\mathbf{X})$.
\end{definition}

\vspace{-0.3cm}
\begin{remark1}
 Eq.~\eqref{eq:equiv} can be reformulated more intuitively as: 
\begin{align}
\hspace{-0.5em}
   F(\cdot|\mathbf{X}) = F\big(T_g(\cdot)|T_g(\mathbf{X})\big), \quad \text{for all } g \in G, \mathbf{X} \in \mathcal{X}.
   \label{eq:equiv2}
\end{align}
\end{remark1}
 Eq.~\eqref{eq:equiv2} indicates that the equivariance is satisfied if for any observation $\mathbf{X}$ and query $ \vec{p}$, the implicit output of $F(\vec{p}|\mathbf{X})$ is \emph{locally} invariant to any  $T_g$ applied jointly to $\mathbf{X}$ and $\vec{p}$ in the implicit model.

\begin{figure*}[t]
\centering 
\vspace{-0.6em}
\includegraphics[width=1.0\linewidth]{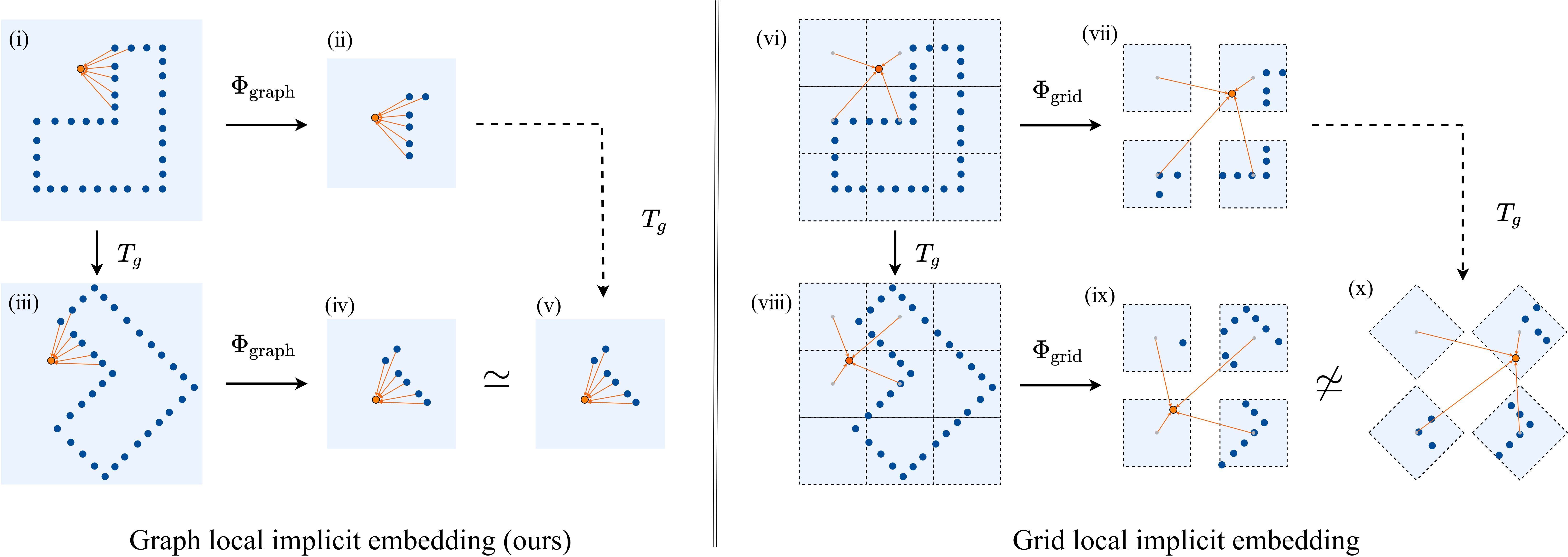}
\vspace{-0.6em}
\caption{
\textbf{Graph ({\it left}) vs. grid ({\it right}) local implicit feature embeddings under rotation.} 
$\Phi_{\text{graph}}$ extracts $k$-NN graphs and applies graph convolutions; $\Phi_{\text{grid}}$  partitions points into regular grids and applies regular convolutions.
{\it Left}:  given a point cloud observation $\mathbf{X}$ ({\it navy}) (\romannumeral1), our method aggregates the local latent feature at any query coordinate $\vec{p}$ ({\it orange}) from a local $k$-NN graph connecting its neighbours in $\mathbf{X}$ (\romannumeral2). Moreover,  when a transformation $T_g$, {\it e.g.} rotation, is applied to the shape, %
(\romannumeral4) the constructed local graph is in the same structure as  (\romannumeral5) applying  $T_g$ to the graph from untransformed data. 
{\it Right}: in contrast,  (\romannumeral6) visualizes discretized grid features. (\romannumeral7) The off-the-grid query location $\vec{p}$ interpolates the neighboring on-grid features. However,  (\romannumeral8) with $T_g$ applied to the raw observation, often (\romannumeral9) the sub-grid point patterns for the local features are different from  (\romannumeral10) applying  $T_g$ to the untransformed local grids. This makes the grid local implicit models sensitive to transformations such as rotation.
}
\vspace{-0.1em}
\label{fig:method1}  
\end{figure*}

\section{Transformation-robust Graph Local Implicit Representations}

Our goal is to design an equivariant implicit function model using local feature embeddings to capture fine  details of the 3D geometry. However, existing grid-based local implicit functions are sensitive to geometric transformations such as rotations, thus not suitable for equivariant implicit representations. To address this limitation, we propose a graph-based local embedding which is robust to geometric transformations.

\vspace{0.3em}
\noindent
\textbf{Background: grid local implicit representations.}
To overcome the limitation of a global latent feature, recent methods, such as ConvONet~\cite{peng2020convolutional} and IF-Net~\cite{chibane2020ifnet}, propose to learn  spatially-varying latent features $\vec{z}_{\vec{p}}  = \Phi_\text{grid}(\vec{p}; \mathbf{X} )$.
The main idea is to partition the 3D space into a grid and compute latent codes locally. %
Specifically, these methods formulate the local latent implicit function on the grid as $F_\text{grid}(\vec{p}| \mathbf{X}) = \Psi(\vec{p}, \vec{z}_{\vec{p}})= \Psi(\vec{p}, \Phi_\text{grid}(\vec{p};\mathbf{X}))$, where the {\it grid local latent extractor} $\Phi_\text{grid}$ is further decomposed as $ \Phi_\text{grid}(\vec{p};\mathbf{X})= \psi_{\text{grid}}\left(\vec{p}, \phi_{\text{grid}}(\mathbf{X}) \right)$. %
The function $\phi_\text{grid}$ is the {\it grid feature encoder}   %
that learns to generate a 2D or 3D grid-based feature tensor $\mathbf{M}$ from the entirety of point observations $\mathbf{X}$. %
For each grid location, point features are aggregated
for all $\vec{x}_i \in \mathbf{X}$  in the corresponding bin.
Convolutional layers are applied to the grid-based $\mathbf{M}$ to capture multi-scale information and maintain translation equivariance. 
Given the local 3D feature tensor $\mathbf{M}$, the  {\it local latent aggregator} $\psi_\text{grid}$ computes local latent feature $ \vec{z}_{\vec{p}}$ on any off-the-grid query coordinate $\vec{p}$ using a simple trilinear interpolation.
We refer to \cite{jiang2020lig,chabra2020deeplocalshapes} for other variants of grid-based representations using purely local information without global cues, while restricted by requiring additional priors during inference such as the normals.
Overall, grid partitioning is not robust to general transformations such as rotations, especially when the sub-grid structure is considered for the resolution-free implicit reconstruction.  %
Fig.~\ref{fig:method1}  ({\it right}) shows an illustration of this limitation, which we will address in our method.

\subsection{Graph-structured local implicit feature embeddings}
\label{sec:graph1}

We propose to use graphs as a non-regular representation, such that our local latent feature function is robust to these transformations and free from feature grid resolutions.
The graph-local implicit function extends the standard form of Eq.~\eqref{eq:global_latent} to 
   \begin{align}
  \vspace{-0.5em}
    F_\text{graph}(\vec{p}| \mathbf{X}) = \Psi(\vec{p}, \vec{z}_{\vec{p}})= \Psi(\vec{p}, \Phi_\text{graph}(\vec{p};\mathbf{X}))= \Psi(\vec{p}, \psi\left(\vec{p}, \phi(\mathbf{X}) \right). \label{eq:graph_f}
 \end{align}
 $\Phi_\text{graph}$ is a deep network that extracts \emph{local} latent features $\vec{z}_{\vec{p}}$ on the graph, composed of two sub-networks $\phi$ and $\psi$. 
The \emph{point feature encoder} $\phi$ maps the point set $ \mathbf{X} = \{ \vec{x}_i\}$ to the associated features  $\{ \vec{h}_i\}$, and is invariant to the sampled query location $\vec{p}$. 
The \emph{graph local latent feature aggregator} $\psi$ propagates the input point feature to the query coordinate $\vec{p}$. 
Unlike grid-based methods, we directly aggregate local information from the point cloud feature  $\{( \vec{x}_i,  \vec{h}_i)\}$ without an intermediate grid feature tensor. 
In particular, we construct a local $k$-nearest neighbor ($k$-NN) graph $(\mathcal{V}, \mathcal{E})$ for every query point $\vec{p}$, where the vertices $\mathcal{V} = \mathbf{X} \cup\{\vec{p}\} $ include the point set elements and the query coordinate.
The edges $ \mathcal{E} = \{(\vec{p}, \vec{x}_i) \}$ are between the query point $\vec{p}$ and its $k$-NN points from the observation point set $ \vec{x}_{i'}  \in \mathcal{N}_k({\vec{p}, \mathbf{X}})$, with $\mathcal{N}_k({\vec{p}, \mathbf{X}})$ denoting the set of the $k$-NN points of $\vec{p}$ from $\mathbf{X}$.
Last, with graph convolutions, we aggregate into the local feature vector $\vec{z}_{\vec{p}}$ the point features of the neighbors of $\vec{p}$.
We adopt a simple spatial graph convolution design in the style of Message Passing Neural Network (MPNN)~\cite{gilmer2017mpnn}, which is widely used for 3D shape analysis~\cite{wang2019dgcnn,hanocka2019meshcnn}. 
For each neighboring point $\vec{x}_{i'}$ from the query $\vec{p}$, messages are passed through a function $\eta$ as a shared two-layer ReLU-MLP, where the inputs are the point features $\vec{h}_{i'}$ and the query coordinate $\vec{p}$ as node feature as well as the displacement vector $\vec{x}_{i'}$ -  $\vec{p}$ as the edge feature, followed by a permutation-invariant aggregation $\text{AGGRE}$ over all neighboring nodes, e.g., max- or mean-pooling:
\begin{align}
  \vspace{-0.5em}
    \vec{z}_{\vec{p}}&=  \aggre_{i'} \eta(\vec{p}, \vec{h}_{i'}, \vec{x}_{i'}  - \vec{p} ).
      \vspace{-0.5em}
    \label{eq:graph_decoder_singlescale}
\end{align}
For each neighboring point as a graph node, the edge function $\eta$ take as inputs, the query coordinate $\vec{p}$, the node point feature $\vec{h}_{i'} $, and the relative position $\vec{x}_{i'}  - \vec{p}$. 

As all the graph connections are relative between vertices, the local latent feature aggregation is robust to transformations like rotations, as illustrated in Fig.~\ref{fig:method1} (\textit{left}).

\begin{figure}[t]
\centering 
\vspace{-0.7em}
\includegraphics[trim={6cm 0 0 0},clip,width=0.65\linewidth]{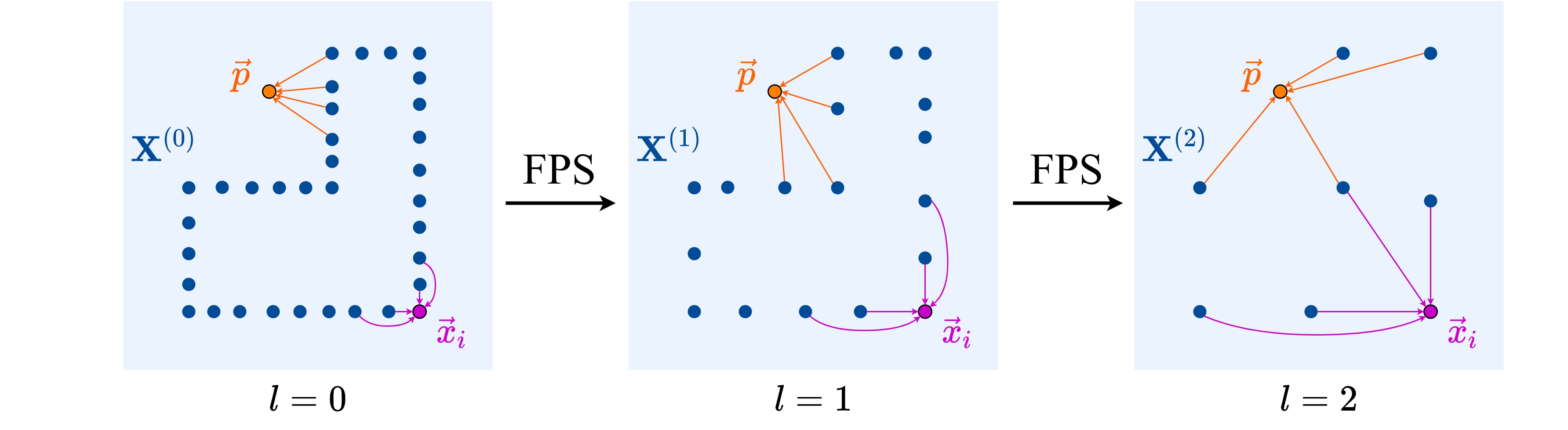}
\caption{ \textbf{Multi-scale design,} with enlarged receptive fields of local $k$-NN graphs for the graph point encoder $\phi$ ({\it orange}) and the graph local latent aggregator $\psi$ (\textit{violet}). 
}
\vspace{-0.6em}
\label{fig:multiscale}  
\end{figure}

\subsection{Learning multi-scale local graph latent features}
\label{sec:graph2}

To capture the context of the 3D geometry at multiple scales, 
both ConvONet~\cite{peng2020convolutional} and IF-Net~\cite{chibane2020ifnet} rely on a convolutional U-Net~\cite{ronneberger2015unet}, with progressively downsampled and then upsampled feature grid resolutions to share neighboring information at different scales. %
Our graph model enables learning at multiple scales by {farthest point sampling} (FPS).
That is, we downsample the point set $\mathbf{X}$ to $\mathbf{X}^{(l)}$ at sampling levels $l=1,\dots,L$ with a progressively smaller cardinality  $|\mathbf{X}^{(l)}| < |\mathbf{X}^{(l-1)}|$, where  $\mathbf{X}^{(0)} = \mathbf{X}$ is the original set.

Moreover, we use a graph encoder for the point encoder $\phi$, instead of PointNet~\cite{qi2017pointnet}.
This way, without involving regular grid convolutions, we can still model local features and facilitate a translation-equivariant encoder, which is beneficial in many scenarios,  especially for learning scene-level implicit surfaces~\cite{peng2020convolutional}. 
Next, we sketch the multi-scale graph point encoder and latent feature aggregator, with Fig.~\ref{fig:multiscale} as a conceptual illustration.

The \textit{graph point encoder} $\phi$ learns point features $\{h^{(l)}_i\}$ for the corresponding points $x_i \in \mathbf{X}^{(l)}$ at each sampling level $l=0\dots, L$.
The encoder starts from the initial sampling level $l=0$ where the input features are the raw coordinates. At each sampling level, a graph convolution is applied to each point to aggregate message from its local $k$-nearest neighbor point features, followed by an FPS operation to the downsampled level $l+1$. 
The graph convolution is similar to that in Eq.~\eqref{eq:graph_decoder_singlescale}, with the point features from both sides of the edge and the relative position as inputs. The graph convolutions and FPS downsampling are applied until the coarsest sampling level $l=L$. Then the point features are sequentially upsampled back from $l=L$ to $l=L-1$, until $l=0$. At each sampling level $l$, the upsampling layer is simply one linear layer followed by ReLU activation. For each point, the input of the upsampling layer is the nearest point feature from the last sampling level $l+1$, and the skip-connected feature of the same point at the same sampling level from the downsampling stage.
Thus far, we obtain multi-scale point features $\{(\vec{x}_i , \vec{h}_i^{(l)})\}$ for $\vec{x}_i \in \mathbf{X}^{(l)}$ at sampling levels $l=0,\dots,L$, as the output from the graph point encoder $\phi$.

For the \textit{graph latent feature aggregator} $ \psi$, at each query coordinate $\vec{p}$, we use graph convolutions to aggregate the $k$-neighboring features $\{ (\vec{x}_i,\vec{h}_i^{(l)})\}$ at different sampling levels $l$, as described in Sec~\ref{sec:graph1}.
The aggregated features  from all sampling levels $l$ are concatenated to yield the local latent vector $\vec{z}_{\vec{p}}$ as output.  The detailed formulations of $ \phi$ and $ \psi$ are provided in the supplementary material.

\section{Equivariant Graph Implicit Functions}

{The local graph structure of the proposed implicit function, with $\mathbf{X} \cup \{\vec{p}\}$ as the set of vertices, is in line with the requirement of equivariance in Sec.~\ref{sec:equiv} and Eq.~\eqref{eq:equiv2}. As a result, the local graph implicit embedding can be used for an equivariant model to achieve theoretically guaranteed generalization to unseen transformations. }
To do this, we further require all the graph layers  in the latent extractor $\Phi_\text{graph}$ to be equivariant in order to obtain equivariant local latent feature. 
In addition, we remove the query coordinate input $\vec{p}$ to ensure the implicit decoder $\Psi$ spatially invariant in Eq.~\eqref{eq:graph_f}, as the local spatial information is already included in the latent $\vec{z}_{\vec{p}}$. 
See Appendix A for details.

We first look into the equivariant layers for 3D rotation group $G = SO(3)$, a difficult case for implicit functions~\cite{davies2020effectiveness}. Then, we discuss how to extend the method to other similarity transformations, such as translation and scaling.

\subsection{Hybrid feature equivariant layers}
Our equivariant layers for the graph convolution operations is inspired by recent methods
 \cite{deng2021vectorneurons,shen2020quaternion} %
 that lift from a scalar  neuron feature $h \in \mathbb{R}$ to a vector  $\vec{v} \in \mathbb{R}^3$ to encode rotation, and the list of 3D vector features $\mathbf{V} = [\vec{v}_1, \vec{v}_2, \dots, \vec{v}_{C_{\mathbf{v}}} ]^\top \in \mathbb{R}^{C_{\mathbf{v}}\times 3}$ that substitutes the scalar features $\vec{h} \in \mathbb{R}^{C_h}$. 
However, using only vector features in the network is non-optimal for both effectiveness and  efficiency, with highly regularized linear layers and computation-demanding nonlinearity projections.

To this end, we extend the method and propose hybrid features $\{\vec{h}, \mathbf{V}\}$, where vector features $ \mathbf{V}$ encode rotation equivariance, and scalar features $\vec{h}$ are rotation-invariant. In practice, hybrid features  show
improved performance and computation efficiency by transferring some learning responsibility to the scalar features through more powerful and efficient standard neural layers.

\begin{figure}[t]
\centering 
\vspace{-0.5em}
\includegraphics[width=0.7\linewidth]{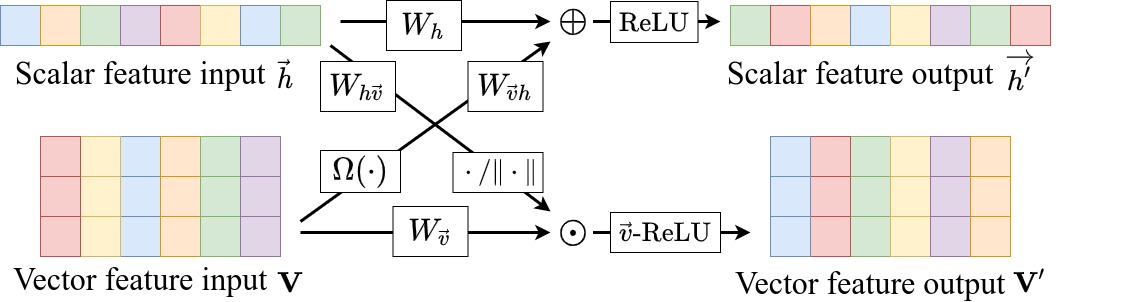}
\caption{\textbf{Hybrid feature equivariant layers.} Visualization of how vector and scalar features share information in linear and nonlinear layers. Vector features go through an invariant function $\Omega$ that is added to the scalar part. Scalar features are transformed  with  normalizing $\cdot / \|\cdot\|$ to scale the vector feature channels. %
}
\vspace{-0.4em}
\label{fig:hn}  
\end{figure}

\vspace{0.3em}
\noindent\textbf{Linear layers. } 
For input hybrid hidden feature $\{\vec{h}, \mathbf{V}\}$ with  $\vec{h} \in \mathbb{R}^{C_h}$, and $\mathbf{V} = [\vec{v}_1, \vec{v}_2, \dots, \vec{v}_{C_{\mathbf{v}}} ]^\top \in \mathbb{R}^{C_{\mathbf{v}}\times 3}$,
we define a set of weight matrices, $\mathbf{W}_{h}\in \mathbb{R}^{C'_h \times C_h}$, $\mathbf{W}_{\vec{v}}\in  \mathbb{R}^{ C'_{\vec{v}} \times  C_{\vec{v}}}$, $\mathbf{W}_{h\vec{v}}\in  \mathbb{R}^{ C'_{\vec{v}} \times C_h}$, and $ \mathbf{W}_{\vec{v}h} \in  \mathbb{R}^{C'_h \times  C_{\vec{v}}} $ for the linear transformation, with information shared between scalar and vector features in the inputs. The resulting output features $\{\vec{h}', \mathbf{V}'\}$ become: %
\begin{align}
         \vspace{-0.3em}
\vec{h}' &= \mathbf{W}_{h}\vec{h} +  \mathbf{W}_{\vec{v}h} \hspace{0.2em}\Omega(\mathbf{V})     \label{eq:hnlinear0}\\
     \mathbf{V}' &= \mathbf{W}_{\vec{v}} \mathbf{V} \odot  \left( \mathbf{W}_{h\vec{v}}\hspace{0.2em}\vec{h} \hspace{0.2em}/\hspace{0.2em} \| \mathbf{W}_{h\vec{v}}\hspace{0.2em}\vec{h}\| \right)
     \label{eq:hnlinear}
     \vspace{-0.3em}
\end{align}
where  $\odot$ is channel-wise multiplication between $\mathbf{W}_{\vec{v}} \mathbf{V} \in \mathbb{R}^{ C'_{\vec{v}} \times 3}$ and $ \mathbf{W}_{h\vec{v}}\vec{h} \in \mathbb{R}^{ C'_{\vec{v}} \times 1}$, and the normalized  transformed scalar feature $\mathbf{W}_{h\vec{v}}\hspace{0.2em}\vec{h} \hspace{0.2em}/\hspace{0.2em} \| \mathbf{W}_{h\vec{v}}\hspace{0.2em}\vec{h}\|$ learns to scale the output vectors in each channel; $\Omega(\cdot)$ is the invariance function that maps the rotation equivariant vector feature $\mathbf{V} \in \mathbb{R}^{ C_{\vec{v}} \times 3}$ to the rotation-invariant scalar feature $\Omega(\mathbf{V}) \in \mathbb{R}^{ C_{\vec{v}}}$, to be added on the output scalar feature. The design of the invariance function $\Omega(\cdot)$ is introduced later in Eq.~\eqref{eq:invariance}.

\vspace{0.3em}
\noindent\textbf{Nonlinearities. }
Nonlinearities apply separately to scalar and vector features. For scalar features, it is simply a $\text{ReLU}(\cdot)$. For vector features, the nonlinearity $\vec{v}\text{-ReLU}(\cdot)$ adopts the design from Vector Neurons \cite{deng2021vectorneurons}:  the vector feature $\vec{v}_c$ at each channel $c$ takes an inner-product with a learnt direction $\vec{q} = [\mathbf{W}_{\vec{q}}\mathbf{V}]^\top \in \mathbb{R}^{3}$ from a linear layer $\mathbf{W}_{\vec{q}} \in \mathbb{R}^{1\times  C_{\vec{v}}}$. If the inner-product is negative, $\vec{v}_c$ is projected to the plane perpendicular to $\vec{q}$.

\begin{equation}
    \hspace{-2em}[\vec{v}\text{-ReLU}(\mathbf{V})]_c = \begin{cases}
             \vec{v}_c & \text{if } \left<\vec{v}_c,  \frac{\vec{q}}{\|\vec{q}\|} \right>\geq 0, \\  
             \vec{v}_c - \left< \vec{v}_c, \frac{\vec{q}}{\|\vec{q}\|} \right> \frac{\vec{q}}{\|\vec{q}\|}. & \text{otherwise}.
             \end{cases}\label{eq:vrelu}
\end{equation}

\vspace{0.5em}
\noindent\textbf{Invariance layer. }
The invariance function $\Omega(\cdot)$  maps the rotation equivariant vector feature $\mathbf{V} \in \mathbb{R}^{ C_{\vec{v}} \times 3}$ to a rotation-invariant scalar feature $\Omega(\mathbf{V}) \in \mathbb{R}^{ C_{\vec{v}}}$ with the same channel dimension $C_{\vec{v}}$. 
At each layer $c=1,\dots,C_{\vec{v}}$, $[\Omega(\mathbf{V})]_c \in \mathbb{R}$ takes the inner product of $\vec{v}_c$ with the channel-averaged direction:
\begin{align}
\vspace{-0.5em}
    [\Omega(\mathbf{V})]_c = \left< \vec{v}_c, \frac{\overline{\vec{v}}}{\|\overline{\vec{v}}\|} \right>,
    \label{eq:invariance}
\vspace{-0.5em}
\end{align}
where  $\overline{\vec{v}} = \frac{1}{C_{\vec{v}}}\sum_{c'=1}^{C_{\vec{v}}} \vec{v}_{c'}$ is the channel-averaged vector feature. One can verify that applying any rotation on the vector feature does not change the inner product.  Our $\Omega$ adopts from \cite{deng2021vectorneurons} with modification. Ours is parameter-free using averaged direction, while \cite{deng2021vectorneurons} learns this vector.

The invariance function is applied in each linear layer in Eq.~\eqref{eq:hnlinear} to share the information between vector and scalar parts of the feature. In addition, at the end of equivariant graph feature network $\Phi_{\text{graph}}$,  $\Omega(\mathbf{V})$ is concatenated with the scalar feature as the final invariant local latent feature ${\vec{z}}_{\vec{p}} =\Omega(\mathbf{V}) \| \vec{h}$ , as to be locally invariant with transformed $\vec{p}$ and $\mathbf{X}$ in line with Eq.~\eqref{eq:equiv2}.

\subsection{Extension to similarity transformations}
While existing vector-based equivariance methods~\cite{deng2021vectorneurons,shen2020quaternion} apply only to the $\text{SO}(3)$ group\footnote{More generally, it is the $\text{O}(3)$ group. Reflection is handled  as well.},  we extend our method to be equivariant to the similarity transformation group that further includes translation and scale transformations as subgroups. 

\vspace{0.3em}
\noindent\textbf{Translation. }
The local graph structure is robust to rotation by design. %
The method can further achieve numerically guaranteed translation equivariance simply by removing the absolute coordinates input  $\vec{p}$ from the graph layers in Eq.~\eqref{eq:graph_decoder_singlescale},  
keeping only the relative positions as the spatial cue. %

\vspace{0.3em}
\noindent\textbf{Scale.}
As vectors hold scale information from their norms, we extend the method for scale equivariance by modifying to normalize the invariance function $\Omega(\mathbf{V}) \leftarrow \Omega(\mathbf{V}) / \|\Omega(\mathbf{V})\|$ based on Eq.~\eqref{eq:invariance}. Likewise, in each layer the scalar features are scale invariant and the vectors are scale equivariant.  %

\section{Experiments}

\begin{table}[t]
\centering
\vspace{-1em}
\caption{\textbf{ShapeNet implicit reconstruction from sparse noisy point clouds.} The grid resolutions are marked for ConvONet and IF-Net. }
\resizebox{0.75\columnwidth}{!}{%
\begin{tabular}{ccccc}
\toprule
 & SO(3) equiv. & Mean IoU $\uparrow$ & Chamfer-$\ell_1$ $\downarrow$ & Normal consist. $\uparrow$  \\
\midrule
ONet & $\times$ &  0.736 & 0.098 & 0.878\\
ConvONet-2D ($3$$\times$$64^2$)& $\times$  & 0.884 & 0.044 & 0.938  \\
ConvONet-3D ($32^3$)& $\times$  &  0.870 & 0.048 & 0.937  \\
IF-Net ($128^3$)& $\times$  & 0.887& 0.042&0.941 \\
GraphONet (ours) & $\times$ & \textbf{0.904} & \textbf{0.038} & \textbf{0.946}\\ \midrule
VN-ONet &   $\checkmark$ & 0.694 & 0.125 & 0.866 \\
E-GraphONet-SO(3) (ours)& $\checkmark$   & {0.890} & 0.041 & 0.936\\
\bottomrule
\end{tabular}
}
\label{tab:shapenet}
\vspace{-0.3em}
\end{table}

\begin{figure}[t]
\centering 
\includegraphics[width=1.02\linewidth]{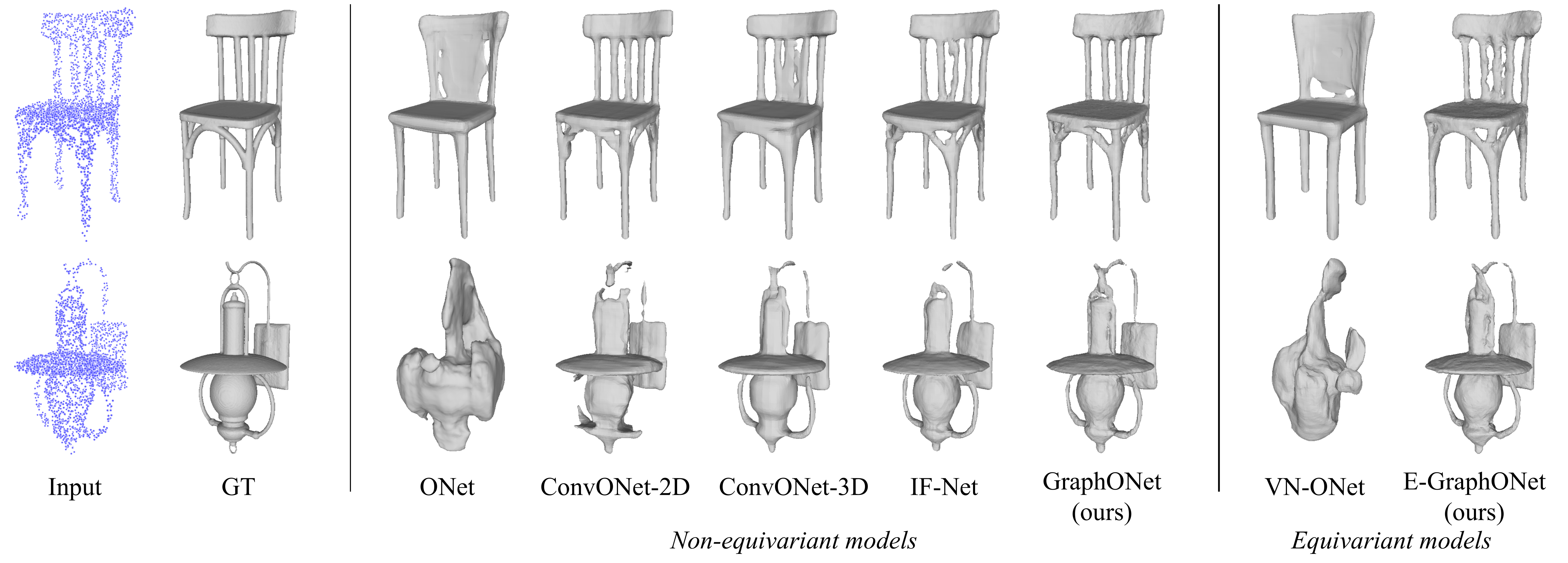}
\vspace{-1.8em}
\caption{\textbf{ShapeNet object reconstruction in canonical space.} GraphONet is among the best performing non-equivariant implicit representation methods; E-GraphONet significantly improves on the existing equivariant method VN-ONet~\cite{deng2021vectorneurons}. 
}
\vspace{-0.3em}
\label{fig:shapenet}  
\end{figure}

We experiment on implicit surface reconstruction from sparse and noisy point observations. %
In addition, we evaluate the implicit reconstruction performance under random transformations of rotation, translation and scaling. 
Our method is referred to as Graph Occupancy networks, or GraphONet, while E-GraphONet is the equivariance model with 3 variants: SO(3), SE(3) and  similarity transformations (Sim.).

 \vspace{0.3em}
\noindent\textbf{Implementation details.}
We implement our method using PyTorch~\cite{paszke2019pytorch}. The number of neighbours in $k$-NN graph is set as 20.
 For the multi-scale graph implicit encoder, we take $L=2$, {\it i.e.},  the point set is downsampled twice with farthest point sampling (FPS) to 20\% and 5\% of the original cardinality respectively. %
 The permutation invariant function $\aggre$ \hspace{-0.3em}adopts mean-pooling for vector features and max-pooling for scalar features.
We use an Adam optimizer~\cite{kingma2014adam} with $\gamma = 10^{-3}$, $\beta_1 = 0.9$ and $\beta_2 = 0.999$. Main experiments are conducted on the {{ShapeNet}}~\cite{chang2015shapenet} dataset with human designed objects, where the train/val/test splits follow prior work~\cite{choy2016universal_correspondence,peng2020convolutional} with 13 categories. Following \cite{peng2020convolutional}, we sample 3000 surface points per shape and apply Gaussian noise with 0.005 standard deviation.
 More details are provided in the supplementary material. 
 
\subsection{Canonical-posed object reconstruction}
We experiment on ShapeNet object reconstruction following the setups in \cite{peng2020convolutional}.
For quantitative evaluation in Table~\ref{tab:shapenet}, we evaluate the IoU, Chamfer-$\ell_1$ distance and normal consistency, following~\cite{mescheder2019occnet,peng2020convolutional}.
The qualitative results are shown in Fig.~\ref{fig:shapenet}. 
Our GraphONet outperforms the state-of-the-arts methods ConvONet~\cite{peng2020convolutional} and IF-Net~\cite{chibane2020ifnet}, as our graph-based method aggregates local feature free of spatial grid resolution and captures better local details. 
IF-Net is better than ConvONet but at large memory cost per batch with $128^3$ resolution grid feature.
For rotation equivariant models, our E-GraphONet significantly outperforms VN-ONet~\cite{deng2021vectorneurons}, benefiting from the graph local features.

\begin{table}[t]
\centering
\caption{\textbf{Implicit surface reconstruction with random rotation, IoU$\uparrow$. }The {\it left} table evaluates non-equivariant methods, and the {\it right} one compares the best performing model with equivariant methods.  \textit{I} denotes canonical pose and  \textit{SO(3)} random rotation.  \textit{I / SO(3)} denotes training with canonical pose and test with random rotation, and so on. { \scriptsize  $^*$: models not re-trained with augmentation due to guaranteed equivariance.} }
\begin{subtable}{0.45\columnwidth}\centering
\resizebox{1\columnwidth}{!}{%
\begin{tabular}{c|ccc}
\toprule
{\it training  / test} &  {\it I / I } & {\it I / $\text{SO}(3)$} & {\it $\text{SO}(3)$ / $\text{SO}(3)$} \\
\midrule
ONet & 0.742 & 0.271 {\scriptsize		 [-0.471]} & 0.592 {\scriptsize [-0.150]} \\
ConvONet-2D &  0.884 & 0.568 {\scriptsize [-0.316]}& 0.791 {\scriptsize [-0.093]} \\
ConvONet-3D & 0.870 & 0.761 {\scriptsize [-0.109]} & 0.838 {\scriptsize [-0.032]} \\
GraphONet (ours) & \textbf{0.904} & \textbf{0.846 {\scriptsize [-0.058]}} & \textbf{0.887 {\scriptsize [-0.017]}} \\
\bottomrule
\end{tabular}
}
\end{subtable}
~
\begin{subtable}{0.52\columnwidth}\centering
\resizebox{1\columnwidth}{!}{%
\begin{tabular}{c|c|ccc}
\toprule
{\it training  / test} &  { equiv.} & {\it I / I } & {\it I / $\text{SO}(3)$} & {\it $\text{SO}(3)$ / $\text{SO}(3)$} \\
\midrule
GraphONet (ours) & $\times$ & \textbf{0.904} & {0.846 {\scriptsize [-0.058]}} & {0.887 {\scriptsize [-0.017]}} \\ \midrule
VN-ONet &  $\text{SO}(3)$ & 0.694 &  0.694 {\scriptsize [-0.000]} & 0.694$^*$ {\scriptsize [-0.000]}\\
E-GraphONet (ours)  &$\text{SO}(3)$& 0.890 & \textbf{0.890 {\scriptsize [-0.000]}}  & \textbf{0.890$^*$  {\scriptsize [-0.000]}} \\\bottomrule 
\end{tabular}
}
\end{subtable}
\label{tab:rot}
\vspace{-0.3em} 
\end{table}
\begin{table}[t]
 \caption{\textbf{Implicit surface reconstruction performance under various types of seen/unseen transformations, mIoU$\uparrow$.} }
\centering
\resizebox{1.02\columnwidth}{!}{%
\begin{tabular}{c|c|c|cc|cc|cc|cc|cc|cc}
\toprule
\multicolumn{1}{c|}{\it Transformation(s)}    &  \multirow{2}{*}{equiv.}      & - %
& \multicolumn{2}{c|}{{\it translation}} & \multicolumn{2}{c|}{{\it scale }} & \multicolumn{2}{c|}{{\it rot. \& transl.}} & \multicolumn{2}{c|}{{\it rot. \&  scale}} & \multicolumn{2}{c|}{{\it transl. \& scale }} & \multicolumn{2}{c}{{\it all }} \\ 
\multicolumn{1}{c|}{\it Training augmentation} &   & - & $\times$  &$\checkmark$&    $\times$   & $\checkmark$  &  $\times$   &  $\checkmark$   &  $\times$   &  $\checkmark$   &   $\times$  &    $\checkmark$ &  $\times$        &   $\checkmark$   \\\midrule
ONet           & $\times$                 &      0.738              &      0.221      & 0.716      &      0.423     &  0.685      &         0.154 &        0.585   &      0.235      &          0.591  &        0.202  &           0.713  &           0.121   &            0.573 \\
ConvONet-2D ($3$$\times$$128^2$)  & $\times^\ddagger$                 &     0.882             &  0.791  &       0.878   &           0.812&     0.850      &     0.532       &     0.771       &          0.542  &    0.789        &          0.723  &      0.838     &       0.481      &      0.728  \\
ConvONet-3D ($64^3$)     & $\times^\ddagger$               &         0.861           &     0.849      &       0.856    &     0.797      & 0.837     &        0.759    &    0.836        &         0.742   &      0.832      &       0.771     &   0.835         &        0.721     &       0.803      \\
GraphONet (ours)   & $\times^\ddagger$              &          \textbf{0.901}          &    \textbf{0.884}       &  \textbf{0.898}      &     0.857      &      \textbf{0.893}     &       0.837     &    \textbf{0.881}        &       0.798     &       {0.880}   &       0.852     &     \textbf{0.888}      &       0.798      &    0.874 \\\midrule
VN-ONet         & $\text{SO}(3)$            &   0.682                 &  0.354         &  0.667         &      0.516     &  0.662         &   0.357         &    0.658        &       0.511     & 0.666           &       0.360     &           0.638 &        0.309     &             0.615\\
E-GraphONet (ours)     & $\text{SO}(3)^\ddagger$      &        0.887        &     0.823       &    0.876 &    0.824  &     0.880   & 0.825           &     0.877       & 0.825           &    \textbf{0.882}       &      0.726       & 0.872     & 0.729 &  0.870     \\
E-GraphONet (ours)     & $\text{SE}(3)$      &        0.884        &     \textbf{0.884}       &    0.884$^*$ &    0.840  &     0.880   & 0.884           &     0.884$^*$       & 0.838           &    {0.880}       &      0.841       & 0.878     & 0.840 &  0.878     \\\midrule
E-GraphONet (ours) & Sim.$^\dagger$   &      {0.882}           &      {0.882}   &    0.882$^*$       &      \textbf{0.882}    &     0.882$^*$      &        \textbf{0.882}    &     0.882$^*$       &    \textbf{0.882}        &        \textbf{0.882$^*$}    &       \textbf{0.882}     &      0.882$^*$      & \textbf{0.882}          &       \textbf{0.882$^*$}     \\\bottomrule
\multicolumn{15}{l}{{\scriptsize $^\dagger$Similarity transformation group. 
$^\ddagger$ The (graph) convolution subnetwork is translation equivariant. 
$^*$with no augmentation due to guaranteed equivariance.} }
\end{tabular}
}

\label{tab:other_transform}
\end{table}

\begin{figure}[t]
\centering 
\vspace{-0.5em}
\includegraphics[width=1.02\linewidth]{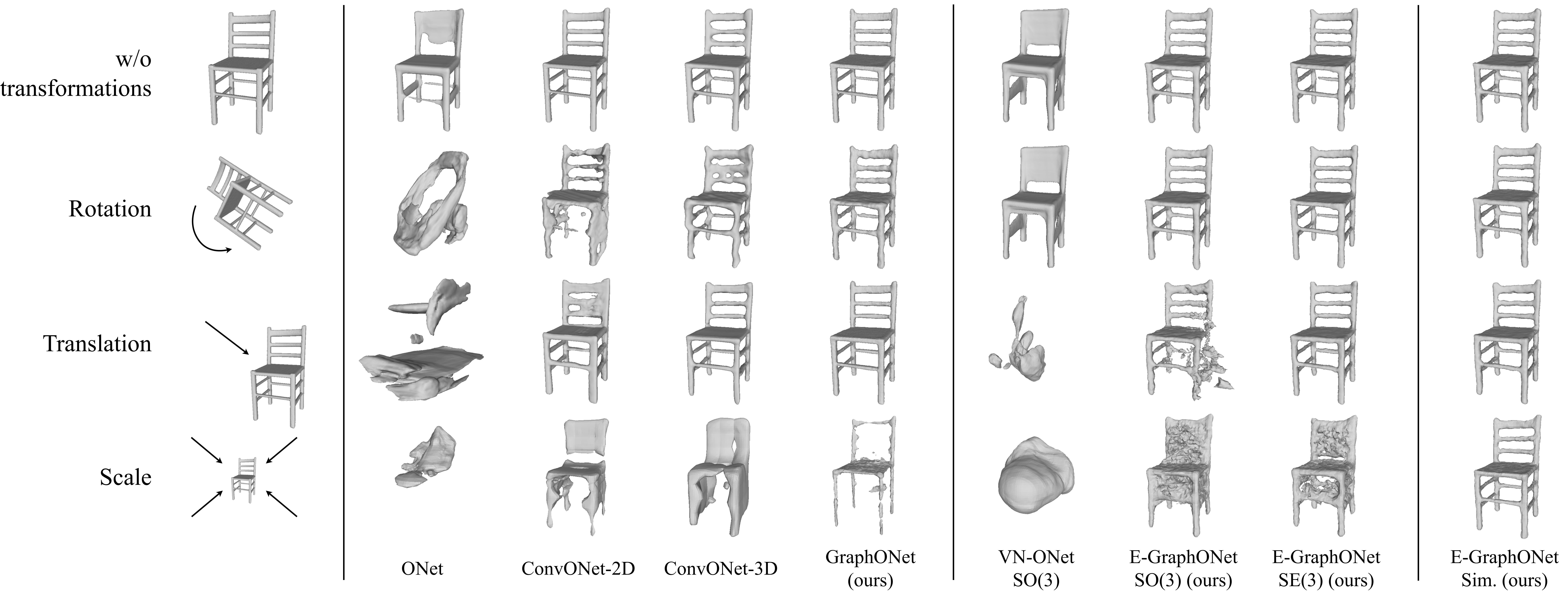}
\vspace{-1.2em}
\caption{\textbf{Implicit surface reconstruction under {\textit{unseen}} transforms.} Visualizing back-transformed shapes. Shapes are scaled by a factor of 0.25, and translation is from the unit cube center to the corner. E-GraphONet-Sim is robust to all similarity transformations.
}
\vspace{-0.6em}
\label{fig:transformations}  
\end{figure}

\subsection{Evaluation under geometric transformations}

\noindent\textbf{Rotation.}
First, we investigate how implicit models perform under random rotations, which is challenging for neural implicits~\cite{davies2020effectiveness}. 
In Table~\ref{tab:rot} left, GraphONet shows the smallest performance drop among all non-equivariant methods under rotations, either with (\textit{SO(3) / SO(3)}) or without augmentation (\textit{I / SO(3)}) during training, since the graph structure is more robust to rotations. 
ConvONet~\cite{peng2020convolutional}-2D is more sensitive than the 3D version, as 3D rotation would lead to highly distinct 2D projections. In Table~\ref{tab:rot} right, E-GraphONet, equipped with equivariant layers, achieves better performance under random rotations, even when the non-equivariant methods are trained with augmentation. It also outperforms the previous equivariant method VN-ONet~\cite{deng2021vectorneurons} by a large margin.

\vspace{0.3em}
\noindent\textbf{Scale, translation and combinations.}
We evaluate how implicit methods perform under various similarity transformations besides SO(3) rotation, including scale, translation and combinations.  
We apply random scales and rotations in a bounded unit cube $[-0.5,0.5]^3$, as assumed by grid methods, and set the canonical scale to be half of the cube.
ConvONet resolutions are doubled to keep the effective resolution.
Random scaling and translation are added under the constraint of the unit bound, with the minimum scaling factor of 0.2. %

From the results in Table~\ref{tab:other_transform}, GraphONet is more robust to transformations than other non-equivariant models. 
For equivariant models, VN-ONet and the SO(3) E-GraphONet models perform poorly on other types of transformations, as they are optimized towards the rotation around origin only. Similarly, the SE(3) E-GraphONet does not generalize to scaling.
Our model with full equivariance performs well on all similarity transformations with numerically the same performance.
Fig~\ref{fig:transformations} shows qualitative examples, where our E-GraphONet-Sim handles all types of unseen similarity transformations.

\subsection{Analysis}
We show some ablation experiments while more results are provided in the Appendix.

\begin{figure}[t]
\noindent
\begin{minipage}[c]{0.5\linewidth}
\vspace{-1em}
\centering
\captionof{table}{\textbf{Parameter- and data-efficiency.} Evaluated on the full test set with 5 runs of 130 randomly sampled training examples. }
\label{tab:ablation_few}  
\vspace{0.2em}
\resizebox{0.87\linewidth}{!}{%
\hspace{-0.7em}\begin{tabular}{ccc}

\toprule
& \#param. & IoU$\uparrow$ \\ \midrule
 \multirow{1}{*}{ConvONet-2D}& \multirow{1}{*}{$2.0 \times 10^6 $} & 0.727 $\pm 0.009$\\
\multirow{1}{*}{ConvONet-3D}& \multirow{1}{*}{$1.1 \times 10^6$}& 0.722 $\pm 0.008$\\
\multirow{1}{*}{GraphONet (ours)}& \multirow{1}{*}{$1.9\times 10^5$} & 0.867 $\pm 0.004$\\
 \multirow{1}{*}{E-GraphONet (ours)}& \multirow{1}{*}{ \textbf{$\mathbf{6.5\times 10^4}$} } & \textbf{0.873 $\pm 0.002$}  \\
 \bottomrule
\end{tabular}
}
\end{minipage}
\hfill
\noindent
\begin{minipage}[c]{0.5\linewidth}
\centering
\hspace{-0.1em}\includegraphics[width=0.78\linewidth]{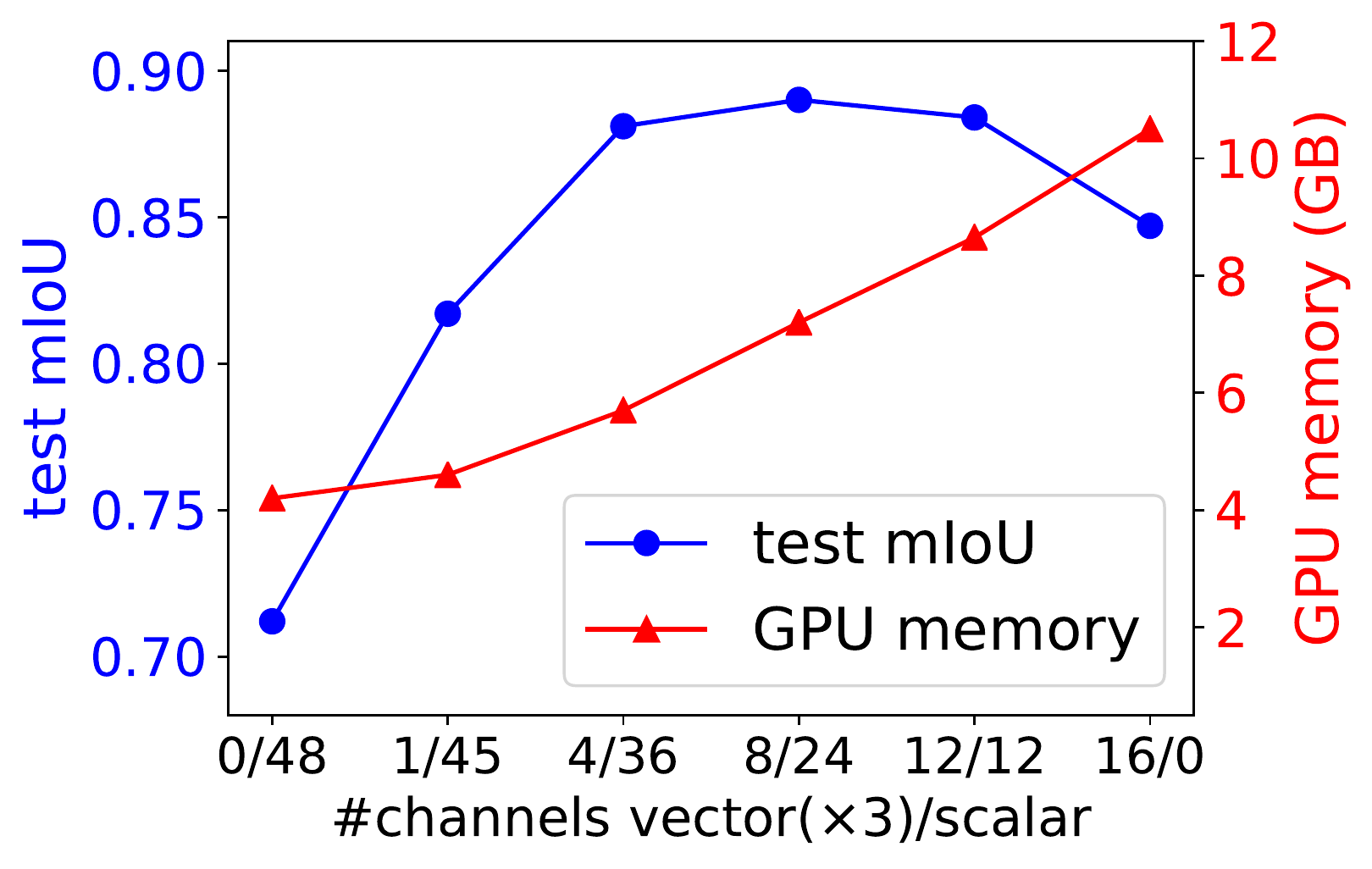}
\vspace{-0.5em}
\captionof{figure}{ \textbf{Ablation on hybrid feature channels.}  Mixed vectors and scalars are more effective and efficient than pure vectors. }
\label{fig:ablation_channel}
\end{minipage}\vspace{-0.6em}
\end{figure}

\vspace{0.3em}
\noindent\textbf{Learning from very few training examples.}
We show that our graph method is both parameter- and data-efficient, and the transform-robust modeling inherently benefits generalization.
As reported in Table~\ref{tab:ablation_few}, we use less than 10\% of the parameters of ConvONet as the graph conv kernel shares parameters for all directions.
We evaluate the test set performance when training on only 130 examples - 10 per class - instead of the full training set size of 30661. While ConvONets fail to achieve good performance, GraphONets does not drop by far from the many-shot results in Table~\ref{tab:shapenet}. 
The E-GraphONet demonstrates even better performance, with more parameter-sharing 
from the equivariance modeling, indicating better power of generalization.

\vspace{0.3em}
\noindent\textbf{Ablation on vector and scalar feature channels.} 
We validate our design of hybrid features by experimenting different ratio of vector and scalar channels. 
We constrain in total 48 effective channels, with one vector channel counted as three scalars. 
In Fig.~\ref{fig:ablation_channel}, using both vectors and scalars with a close-to-equal ratio of effective channels obtains higher performance with less memory cost than using pure vectors, i.e., in the Vector Neurons~\cite{deng2021vectorneurons}.
This indicates the expressive power of the scalar neuron functions. {We provide additional results for non-graph-based equivariant models in the Appendix.}

\subsection{Scene-level reconstructions}

In addition to the ability of handling object shape modeling under transformations, 
our graph implicit functions also scale to scene-level reconstruction. We experiment on two datasets: (\romannumeral1) \textit{{Synthetic Rooms}}~\cite{peng2020convolutional}, a dataset provided by \cite{peng2020convolutional}, with rooms constructed with walls, floors, and  
ShapeNet objects from five classes: chair, sofa, lamp, cabinet and table. 
(\romannumeral2) \textit{{ScanNet}}~\cite{dai2017scannet}, a dataset of RGB-D scans of real-world rooms 
for testing synthetic-to-real transfer performance.

\begin{wraptable}{r}{55mm}
\vspace{-2.4em}
\caption{\textbf{Indoor scene reconstructions.} }
\resizebox{0.5\columnwidth}{!}{%
\begin{tabular}{c|ccc|c}
\toprule
 \multicolumn{1}{c|}{\textit{Dataset}} &    \multicolumn{3}{c|}{\textbf{Synthetic room}} &  \textbf{ScanNet}\\
& IoU$\uparrow$ & Chamfer$\downarrow$ & Normal$\uparrow$  &  Chamfer$\downarrow$ \\
\midrule
ONet & 0.514  & 0.135 & 0.856 & 0.546\\
ConvONet-2D  ($3$$\times$$128^2$) &   0.802 & 0.038 & 0.934 & 0.162\\
ConvONet-3D ($64^3$) &  0.847 & 0.035 & 0.943 & 0.067\\
GraphONet (ours) &  \textbf{0.883} & \textbf{0.032} & \textbf{0.944}& \textbf{0.061} \\ \midrule
E-GraphONet (ours) & 0.851 & 0.035 & 0.934&   0.069\\
\bottomrule
\end{tabular}
}
\label{tab:synthetic_room}
\vspace{-2em}
\end{wraptable}

\begin{figure}[t]
\centering 
\vspace{-0.9em}
\includegraphics[width=0.47\linewidth]{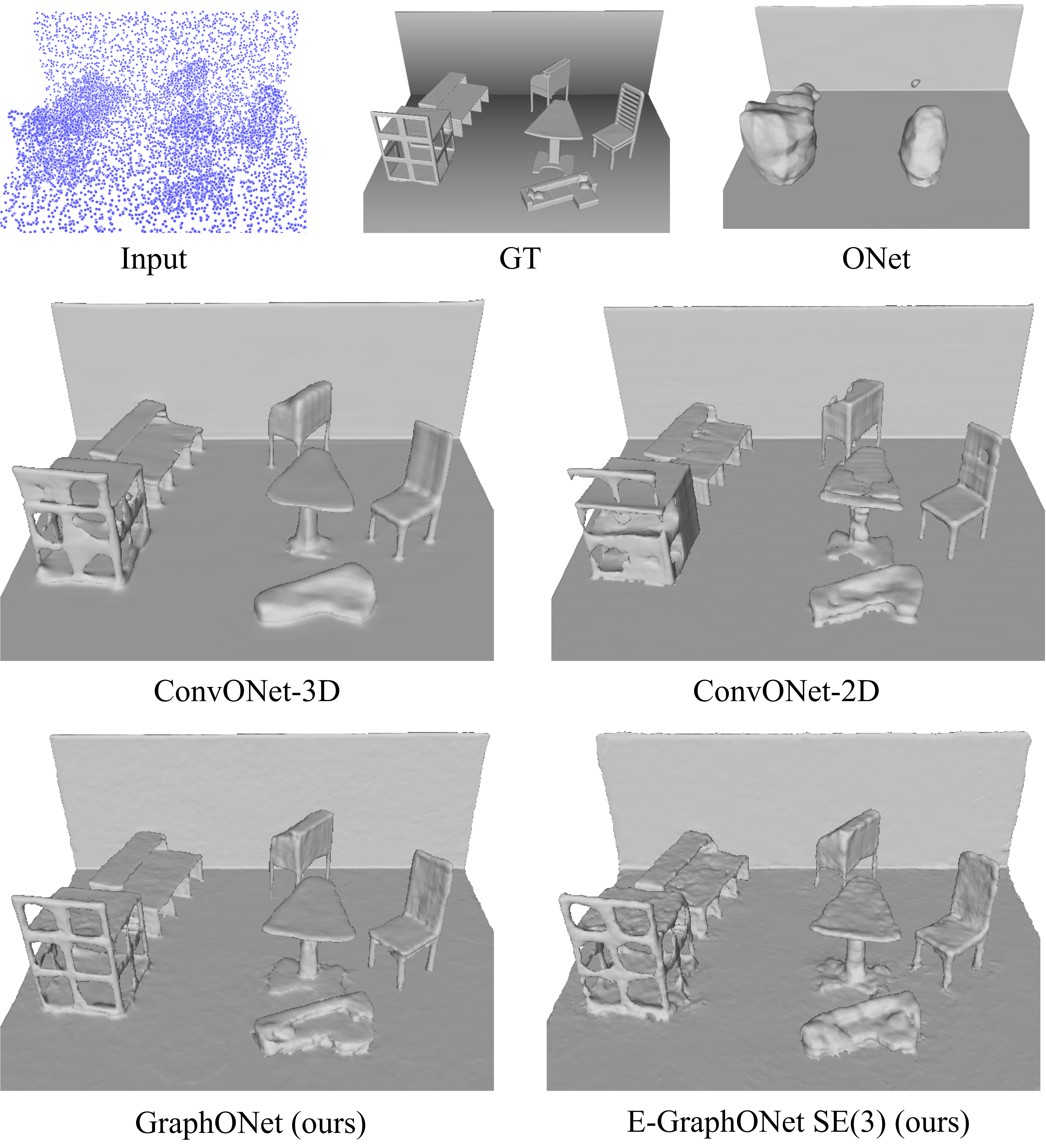}
\hfill
\includegraphics[width=0.52\linewidth]{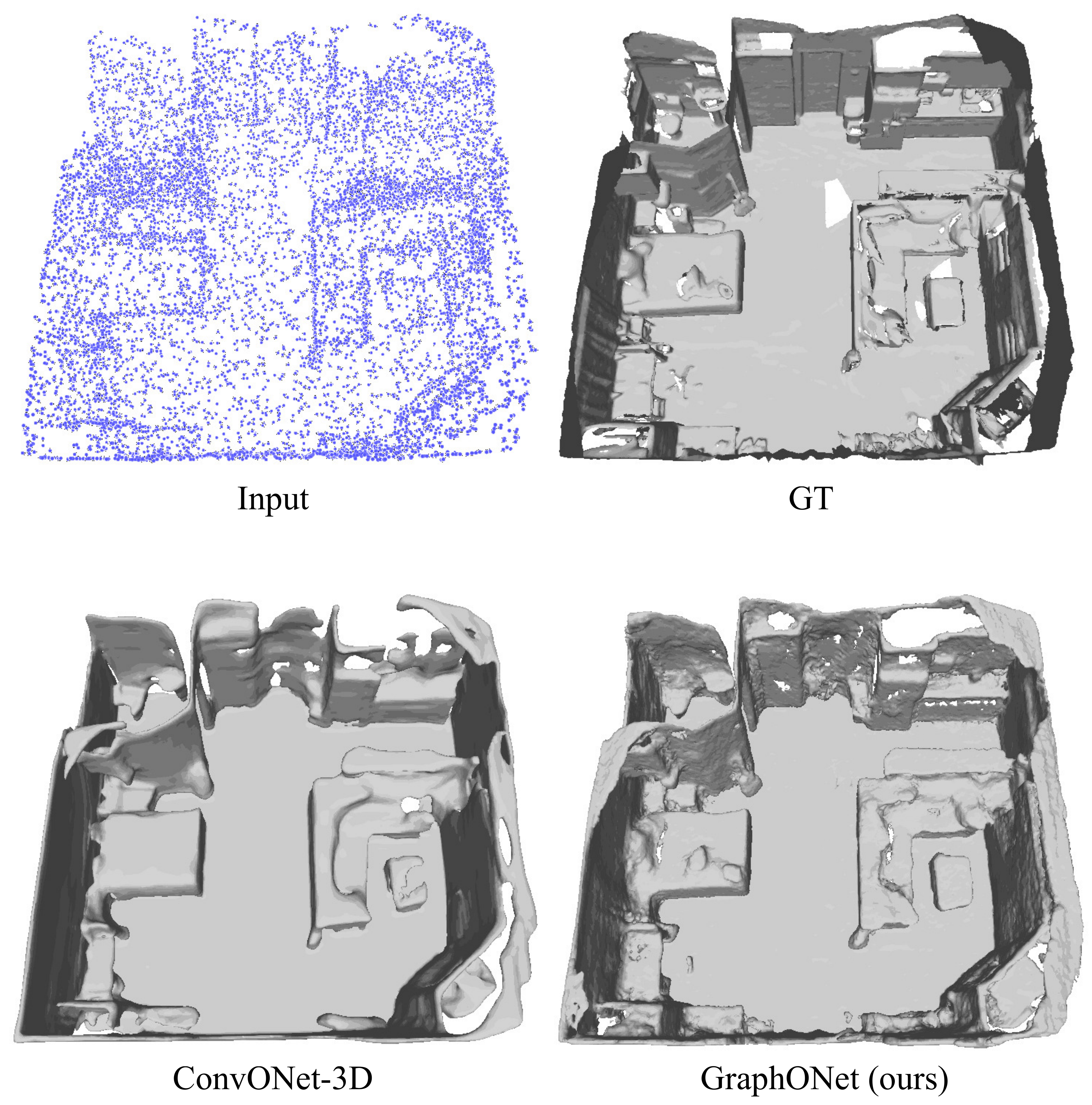}
\vspace{-0.5em}
\caption{ \textbf{Indoor scene reconstructions on Synthetic rooms (\textit{left}) and ScanNet (\textit{right}).} Our GraphONet produces shaper edges and better finer details.} 
\vspace{-0.9em}
\label{fig:rooms}  
\end{figure}

We train and evaluate our model
on Synthetic room dataset~\cite{peng2020convolutional} using 10,000 sampled points as input. 
The quantitative results are shown in Table~\ref{tab:synthetic_room} and qualitative results in Fig.~\ref{fig:rooms} (left). 
Our GraphONet performs better than ConvONets~\cite{peng2020convolutional} at recovering  detailed structures. We evaluate  the SE(3) variant of our equivariance model, and it performs generally well, but less smooth at flat regions. 
In addition, we  evaluate the model transfer ability of our method on ScanNet~\cite{dai2017scannet} with the model trained on synthetic data, for which we report the Chamfer measure in Table~\ref{tab:synthetic_room}. Our GraphONet ourperforms other methods. Fig.~\ref{fig:rooms} (right) shows a qualitative example of our reconstructions.

\subsection{Limitations and future work}

A limitation that comes with equivariance is that, by constraining the model complexity to conform to equivariant designs, expressive power may be affected as well.
As a consequence, accuracy drops in stylized settings and datasets. 
In particular, we observe that the shapes generated from equivariance models are usually less smooth than non-equivariant methods. 
We argue that the restricted power of equivariance models limits the ability to identify the denoised geometry from the noisy point observations, while at the same time, the equivariant model are designed to avoid leveraging the prior of the flat planar structures. 
To this end, relevant future directions include exploring more powerful equivariant models, %
or incorporating filtering techniques for implicit fields~\cite{vladlen2021}.

\section{Conclusion}

In this paper, we introduce graph implicit functions, which learn local latent features from $k$-NN graph on sparse point set observations, enabling reconstruction of 3D shapes and scenes in fine detail. 
By nature of graphs and in contrast to regular grid representations, the proposed graph representations are robust to geometric transformations. %
What is more, we extend the proposed graph implicit functions with hybrid feature equivariant layers, thus guarantee theoretical equivariance under various similarity transformations, including rotations, translations and scales, and obtain models that generalize to arbitrary and unseen transformations. %

\vspace{0.5em}\noindent\textbf{Acknowledgement} This research was supported in part
by SAVI/MediFor project, ERC Starting Grant Scan2CAD (804724), EPSRC programme
grant Visual AI EP/T028572/1 and National Research Foundation, Singapore under its AI Singapore Programme (AISG Award No: AISG2-RP-2020-016). We thank Angela Dai for video voice over.

%
%
\bibliographystyle{splncs04}
\bibliography{00-main.bib}

\clearpage
\newpage

\appendix
\addcontentsline{toc}{section}{Appendices}

\section{Discussion and proof of the equivariance model}

We discuss the equivariance properties of proposed graph equivariant implicit functions and present the mathematical proof for the equivariant layers.

\subsection{Relevance of  layer equivariance and model equivariance}
 We review the definition of the equivariance of the implicit function model in  Section~\ref{sec:equiv}. In Eq.~\eqref{eq:equiv2}, we show that  equivariance is satisfied if the implicit function is locally invariant to any  $T_g$ applied jointly to the observation $\mathbf{X}$ and the query $\vec{p}$, for any  $\mathbf{X}$ and $ \vec{p}$. Invariance is a special case of equivariance, where the transformation $T_g^*$ on the output domain is the identity function, i.e., the output is invariant regardless of the input transformation $T_g$.

In this section, we clarify the use of layer equivariance in creating the equivariant implicit function model.
As in Eq.~\eqref{eq:graph_f}, the graph implicit function model $F_\text{graph}$ is composed of the graph latent feature extractor $\Phi_\text{graph}$ and the implicit decoder $\Psi$,  where  $\Phi_\text{graph}$ can be further decomposed to the point encoder $\phi$ and the graph local latent aggregator $\psi$. 
We integrate the equivariant layers in $\phi$ and $\psi$ that process the input point set $\mathbf{X}$ and the queries $\vec{p}$. 
As shown by the literature, the composition of two equivariant functions is also an equivariant function~\cite{thomas2018tfn}. Thus, the graph local feature extractor  $\Phi_\text{graph}$ that stacks sequential equivariant graph layers in  $\phi$ and $\psi$ is equivariant. Note that all the operations other than the graph convolutions involved in  $\phi$ and $\psi$, such as the  $k$-NN graph extration and the farthest-point-sampling for the multi-scale feature, are equivariant to the similarity transformations as well.
At the end of all the equivariant graph layers in $\phi$ and $\psi$, we apply the 
invariance function $\Omega$ to obtain the locally invariant latent feature. 
The implicit decoder $\Psi$, which processes the invariant local latent feature and predicts the occupancy probability, is simply a standard ReLU-MLP as the non-equivariant implicit models. We do not include the query coordinate input $\vec{p}$ in the implicit decoder $\Psi$ in order to satisfy translation equivariance, while the local latent feature already contains the position information. Since the local latent feature is locally invariant, the output is locally invariant as well, which satisfies Eq.~\eqref{eq:equiv2}. Thus far, we have shown how the equivariant graph layers help to build the equivariant implicit function model.

\subsection{Proof of translation equivariance}
We  discuss the translation equivariance property in separate from other transformation groups. Because the translation equivariance property in our method is from the use of the local graph structure, while the equivariant properties of  other  similarity transformations, including rotations, reflections and scaling, rely on the hybrid features, especially the vector part. Each of the graph layers are locally invariant to the continuous translation group.

For any input points and queries, we discard the absolute global coordinate inputs, and manipulate on the relative positions within a local graph structure in all graph layers in $\phi$ and $\psi$. 
We consider an edge connects an input point  $\vec{x}_{i'}$ and  a query $\vec{p}$ in an equivariant graph layer in the graph latent aggregator $\psi$, then the input vector feature is $\vec{x}_{i'} - \vec{p}$. if a translation vector $\vec{t}$ is applied on all the points, then the input feature becomes 
\begin{equation}
    (\vec{x}_{i'} - \vec{t}) - (\vec{p} - \vec{t}) = \vec{x}_{i'} - \vec{p}.
\end{equation}
Thus, the input is invariant to the translation $\vec{t}$, so is the output of the graph layer. And similarly for the graph layers in the point encoder $\phi$. 
Therefore, the whole graph function is translation-invariant for local predictions from any 3D point inputs, which means that our graph implicit function, local to each of the query locations, is translation equivariant. %

\subsection{Proof of rotation, scaling and reflection equivariance}
Next, we show the equivariance properties on other transformations including rotations,  reflections and scaling.
We assume that the scalar feature $\vec{h}$ is invariant to these transformations, while the vector feature $\mathbf{V}$ is equivariant, before the invariance function $\Omega$ is applied. Formally, we consider an arbitrary orthogonal matrix $\mathbf{Q}\in \mathbb{R}^{3\times3}$ encoding a 3D rotation with a possible reflection, and an arbitrary positive scalar $s\in \mathbb{R}^+$ for the scale transformation applied on the features. The scalar and vector features  $\vec{h}$ and  $\mathbf{V}$ are then transformed into $s\mathbf{VQ}^\top$ and $\vec{h}$ respectively. Note here the orthogonal matrix applying to a stack of vector features $\mathbf{V} = [\vec{v}_1, \vec{v}_2, \dots, \vec{v}_{C_{\mathbf{v}}} ]^\top \in \mathbb{R}^{C_{\mathbf{v}}\times 3}$ returns $(\mathbf{QV}^\top)^\top = \mathbf{VQ}^\top$. We study how the output of each layer changes with the change of the inputs.

\vspace{0.3em}\noindent\textbf{Invariance layer}
We first show that the invariance function works for any $3\times 3$ orthogonal matrix $\mathbf{Q}\in \mathbb{R}^{3\times3}$ encoding 3D rotation and reflection and any random scaling factor $s\in \mathbb{R}$, such that $\Omega(s\mathbf{VQ^\top}) = \Omega(\mathbf{V})$:
\begin{equation}
\begin{aligned}
    \Omega(s\mathbf{VQ}^\top) & =  \frac{\left< s\mathbf{VQ}^\top, \frac{s\mathbf{Q}\overline{\vec{v}}}{\|s\mathbf{Q}\overline{\vec{v}}\|} \right>} { \left\lVert \left< s\mathbf{VQ}^\top, \frac{s\mathbf{Q}\overline{\vec{v}}}{\|s\mathbf{Q}\overline{\vec{v}}\|} \right> \right\rVert} 
      =  \frac{s\left< \mathbf{VQ}^\top, \frac{\mathbf{Q}\overline{\vec{v}}}{\|\overline{\vec{v}}\|} \right>} {s \left\lVert \left< \mathbf{VQ}^\top, \frac{\mathbf{Q}\overline{\vec{v}}}{\|\overline{\vec{v}}\|} \right> \right\rVert} \\
      & =  \frac{\left< \mathbf{Q^{-1} QV}, \frac{\overline{\vec{v}}}{\|\overline{\vec{v}}\|} \right>} { \left\lVert \left< \mathbf{Q^{-1} QV}, \frac{\overline{\vec{v}}}{\|\overline{\vec{v}}\|} \right> \right\rVert} 
         =  \frac{\left< \mathbf{IV}, \frac{\overline{\vec{v}}}{\|\overline{\vec{v}}\|} \right>} { \left\lVert \left< \mathbf{IV}, \frac{\overline{\vec{v}}}{\|\overline{\vec{v}}\|} \right> \right\rVert} \\
                & =  \frac{\left< \mathbf{V}, \frac{\overline{\vec{v}}}{\|\overline{\vec{v}}\|} \right>} { \left\lVert \left< \mathbf{V}, \frac{\overline{\vec{v}}}{\|\overline{\vec{v}}\|} \right> \right\rVert} \\
                & = \Omega(\mathbf{V}),
\label{eq:invariance_proof}
\end{aligned}
\end{equation}
where in the second row of Eq.~\eqref{eq:invariance_proof}, the orthogonal matrices $\mathbf{Q}$ are cancelled out in the inner product.
Here we adopt a slightly abused notation to have the inner product between a stack of vector features  $\mathbf{V} = [\vec{v}_1, \vec{v}_2, \dots, \vec{v}_{C_{\mathbf{v}}} ]^\top$ and the average vector $\overline{\vec{v}}$, such that $\left<\mathbf{V}, \overline{\vec{v}}\right> = [\left<\vec{v}_1, \overline{\vec{v}}\right>, \left<\vec{v}_2, \overline{\vec{v}}\right>, \dots, \left< \vec{v}_{C_{\mathbf{v}}}, \overline{\vec{v}}\right> ]^\top$. %
 
\vspace{0.3em}\noindent\textbf{Linear layer}
Next, we show that our hybrid feature linear layer is equivariant to the rotation, reflection, and scaling transformations, encoded by arbitrary $\mathbf{Q}$ and $s$. %
 Without these transformations, we consider $\{\vec{h}', \mathbf{V}' \}$ as the outputs for $\{\vec{h}, \mathbf{V} \}$ from the layer denoted by $f$, i.e., $\{\vec{h}', \mathbf{V}' \} = f\left(\{\vec{h}, \mathbf{V}\}\right)$; while under the transformations encoded by $\mathbf{Q}$ and $s$, we denote the outputs as $\{ \vec{h}'', \mathbf{V}''\} = f\left(\{  \vec{h}, s\mathbf{V}\mathbf{Q}^\top \}\right)$. 
For the equivariance of $f$, we need to show that:
\begin{align}
     \vec{h}'' = \vec{h}', \text{ and } \mathbf{V}'' = s\mathbf{V}'\mathbf{Q}^\top,
\end{align}
in which we assume that the vector feature $\mathbf{V}$ is equivariant for rotations, reflections and scaling,  while the scalar feature $\vec{h}$ is invariant to these transformations. 

For the scalar feature output in %
Eq.~\eqref{eq:hnlinear0},
one can simply verify the invariance 
\begin{equation}
    \begin{aligned}
    \vec{h}'' &= \mathbf{W}_{h}\vec{h} +  \mathbf{W}_{\vec{v}h} \hspace{0.2em}\Omega(s\mathbf{VQ}^\top) \\
    &= \mathbf{W}_{h}\vec{h} +  \mathbf{W}_{\vec{v}h} \hspace{0.2em}\Omega(\mathbf{V}) \\
    & = \vec{h}'.
    \end{aligned}
\end{equation}
Here the invariance function returns 
\begin{equation}
    \Omega(s\mathbf{VQ}^\top) = \Omega(\mathbf{V}),
\end{equation}
as shown in Eq.~\eqref{eq:invariance_proof}. For the vector feature output in Eq.~\eqref{eq:hnlinear0}, %
\begin{equation}
    \begin{aligned}
    \mathbf{V}'' &= \mathbf{W}_{\vec{v}} \hspace{0.1em} (s\mathbf{VQ}^\top) \odot  \left( \mathbf{W}_{h\vec{v}}\hspace{0.2em}\vec{h} \hspace{0.2em}/\hspace{0.2em} \| \mathbf{W}_{h\vec{v}}\hspace{0.2em}\vec{h}\| \right) \\
    &= s\left( \mathbf{W}_{\vec{v}} \mathbf{V} \odot  \left( \mathbf{W}_{h\vec{v}}\hspace{0.2em}\vec{h} \hspace{0.2em}/\hspace{0.2em} \| \mathbf{W}_{h\vec{v}}\hspace{0.2em}\vec{h}\| \right) \right) \mathbf{Q}^\top\\
    &= s\mathbf{V'Q}^\top
    \end{aligned}
\end{equation}
Thus far we have shown the equivariance of the linear layers.

\vspace{0.3em}\noindent\textbf{Non-linearity}
For the non-linearity, the scalar features take simple ReLU activation, hence the invariance is easily ensured as no equivariant vector feature is involved for the output scalar feature. 

For the vector non-linearity $\vec{v}\text{-ReLU}$ in Eq.~\eqref{eq:vrelu}, %
we first reason that the transformations $\mathbf{Q}$ and $s$ does not influence whether the vector feature $\vec{v}_c$ at each channel $c$ falls in the positive or the negative part of the piecewise non-linearity. With the untransformed feature $\vec{v}_c$,  the positive case is judged by $\left<\vec{v}_c,  \frac{\vec{q}}{\|\vec{q}\|} \right>\geq 0$; while with the transformed feature vector $s\mathbf{Q}\vec{v}_c$,  the learned direction vector $\vec{q}$ is transformed accordingly into $s\mathbf{Q}\vec{q}$. Then, the condition of the positive case becomes $ \left<s\mathbf{Q}\vec{v}_c,  \frac{s\mathbf{Q}\vec{q}}{\|s\mathbf{Q}\vec{q}\|} \right>\geq 0$, which is equivalent to the condition without the transformations, as the orthogonal matrices $\mathbf{Q}$ are cancelled out and the positive scaling factor $s$ does not change the sign. The same for the negative case.

Next, we show the equivariance in both positive and negative cases of the non-linearity in Eq.~\eqref{eq:vrelu}. %
When transformations $\mathbf{Q}$ and $s$ are applied, in the positive case,
\begin{equation}
    \begin{aligned}
    [\vec{v}\text{-ReLU}(s\mathbf{V}\mathbf{Q}^\top)]_c& =
             s\mathbf{Q}\vec{v}_c\\
             &=  s\mathbf{Q}[\vec{v}\text{-ReLU}(\mathbf{V})]_c;
    \end{aligned}
\end{equation}
while in the negative case, 
\begin{equation}
    \begin{aligned}
    [\vec{v}\text{-ReLU}(s\mathbf{V}\mathbf{Q}^\top)]_c &= s\mathbf{Q}\vec{v}_c - \left< s\mathbf{Q} \vec{v}_c, \frac{s\mathbf{Q}\vec{q}}{\|s\mathbf{Q}\vec{q}\|} \right> \frac{s\mathbf{Q}\vec{q}}{\|s\mathbf{Q}\vec{q}\|} \\
    &= s\mathbf{Q}\left( \vec{v}_c - \left< \vec{v}_c, \frac{\vec{q}}{\|\vec{q}\|} \right> \frac{\vec{q}}{\|\vec{q}\|} \right) \\
    &=  s\mathbf{Q}[\vec{v}\text{-ReLU}(\mathbf{V})]_c
    \end{aligned}
\end{equation}
Thus, 
\begin{align}
     \vec{v}\text{-ReLU}(s\mathbf{V}\mathbf{Q}^\top) & = s\left(\vec{v}\text{-ReLU}(\mathbf{V})\right)\mathbf{Q}^\top, \quad \text{or}\\
     \mathbf{V}'' &=  s\mathbf{V'Q}^\top
\end{align}
 has been proven in both the positive and the negative cases of the vector ReLU function, indicating the equivariance of the non-linear layer with regard to rotation, reflection and scaling.

\section{Detailed formulations of graph fuctions in multiple scales}
We provide the detailed formulation of the layers in the multi-scale graph point encoder $\phi$ and the multi-scale graph latent feature decoder $\psi$ in Sec~\ref{sec:graph2}. Here we show the formulations with only the scalar features for the non-equivariant graph implicit model. All these formulations  can be easily adapted to the equivariant model by replacing the scalar features to the hybrid features of both scalars and vectors.

\vspace{0.3em}
\noindent\textbf{Graph point encoder.} 
The point encoder $\phi$ is composed of graph convolution layers in the downsampling stage, starting from $l=0$ to $l=L$, followed by the upsampling layers from $l=L-1$ back to $l=0$.

In each graph convolution layer with the sampled point set $\mathbf{X}^{(l)}$ by farthest-point-sampling in the downsampling stage, we obtain the hidden feature $\vec{h}_i^{(l)}$ for any sampled point at this level $\vec{x}_i \in \mathbf{X}^{(l)}$. To do this, we use graph convolution to aggregate information from the $k$-nearest neighbor points of $\vec{x}_i$, denoted as $\vec{x}_{i'}$. The information from the neighboring points to be aggregated is the hidden feature from the previous layer $\vec{h}_{i'}^{(l-1)}$.
Within the local $k$-NN graph structure, messages are passed through  $ \eta^{(l)}_{\downarrow}$, hence concatenating the inputs for a shared two-layer ReLU-MLP, and a permutation-invariant aggregation    function $\text{AGGRE}$; e.g., max- or mean-pooling operator: 
\begin{align}
\hspace{-1.8em}
   \vec{h}_i^{(l)} = \aggre_{i'} \eta^{(l)}_{\downarrow}(\vec{h}_{i}^{(l-1)}, \vec{h}_{i'}^{(l-1)}, \vec{x}_{i'} - \vec{x}_{i}).
   \label{eq:graph_downsample}
\end{align}
Note that there is an exception in Eq.~\ref{eq:graph_downsample} with $l=0$, where the input features are the raw coordinates.

In each upsampling layer, the point feature  $\vec{h}_i^{(l)}$ for any $\vec{x}_i \in \mathbf{X}^{(l)}$ at a finer sampling level $l$ takes information from $\vec{h}_{i'}^{(l+1)}$, the hidden feature associated to $\vec{x}_i$'s 1-nearest neighbor point  $\vec{x}_{i'}$ from the sampled point set $ \mathbf{X}^{(l+1)}$ from the  previous sampling level. In addition, we skip-connect  $\vec{h}_i^{(l)}$, the feature at the same level from the downsampling stage, which is akin to the U-Net structure in grid-based methods:
\begin{align}
\vspace{-1em}
\hspace{-1em}
     \vec{h}_i^{(l)} \leftarrow   \eta^{(l)}_{\uparrow}(\vec{h}_{i'}^{(l+1)}, \vec{h}_i^{(l)}), %
     \vspace{-1em}
\end{align}
where  $\eta^{(l)}_{\uparrow}$ is a linear layer with a ReLU activation for the concatenation of inputs. %

\vspace{0.3em}
\noindent\textbf{Graph local latent feature aggregator.}
Given the query coordinate $\vec{p}$, we use graph convolutions to aggregate the $k$-neighboring features $\{ (\vec{x}_{i'},\vec{h}_{i'}^{(l)})\}$ at different sampling levels $l$. %
The aggregated features $\vec{z}_{\vec{p}}^{(l)}$ from all sampling levels $l$ are concatenated to yield the local latent vector $\vec{z}_{\vec{p}}$ as output:
\begin{align}
    \vec{z}_{\vec{p}} = \concat_{l=0}^{L} \hspace{0.3em} &\vec{z}_{\vec{p}}^{(l)}, \quad \text{where} \\
     \vec{z}_{\vec{p}}^{(l)} = \aggre_{i'} &\eta^{(l)}(\vec{p}, \vec{h}_{i'}^{(l)}, \vec{x}_{i'}^{(l)}  - \vec{p} ),
    \label{eq:graph_decoder_multiscale}
\end{align}
Likewise, $\eta^{(l)}$ is a two-layer ReLU-MLP for the concatenated inputs, and $\|$ denotes concatenation over sampling levels. %

\section{Implementation details}

We use PyTorch~\cite{paszke2019pytorch} to implement our method and run experiments on a single NVIDIA GeForce GTX 1080 Ti GPU. We train the network using the Adam optimizer~\cite{kingma2014adam} with the initial learning rate is set as $10^{-3}$ for fast convergence for 200K iterations, followed by a finetuning of 100K iterations with the learning rate $10^{-4}$. Other hyperparameters and initializations follow the default setups in PyTorch. 

The number of neighbors in $k$-NN graphs is set as $k=20$ for all the graph convolution layers. %
 For the multi-scale graph structure,  the point set is downsampled twice with farthest point sampling (FPS) to 20\% and 5\% of the original cardinality respectively. 
  The permutation invariant function $\aggre\hspace{-0.2em}$ is a mean-pooling aggregation for vector features and a max-pooling for scalar features. Empirically, we find that using vector max-pooling function as in \cite{deng2021vectorneurons} generates artifacts in the qualitative results, so we simply take the average of the vector features.
For the non-equivariant GraphONet, the number of output feature channels is set as 64 for all the layers in the graph latent feature extractor function $\Phi$. For the equivariance model E-GraphONets with hybrid features, the number of output channels for the vector features is 8, and 32 for scalar features. For the input geometric features, the 3D coordinates or the relative position are considered as 3 channels for the GraphONet, or 1 vector channel and 0 scalar channel for the equivariant layer. For both equivariant and non-equivariant models, the implicit decoder $ F$ is the same as that in the ConvONet~\cite{peng2020convolutional}, which is a light-weight ReLU-MLP architechture with skip-connections.

\section{Additional experiments and results}

\subsection{Vector vs. scalar channels in hybrid feature equivariant layers.}
We extend the ablation experiments on hybrid feature channels in Fig.~\ref{fig:ablation_channel} of the main paper. Here we show that  our hybrid feature paradigm  benefits different architectures and tasks. For implicit surface reconstruction, we evaluate the equivariant implicit model without a graph embedding. We 
follow the VN-ONet architecture and the implementation details from \cite{deng2021vectorneurons},
 and use hybrid layers instead of pure vector neuron layers. In addition,
 we evaluate point cloud classification on the ModelNet40 dataset. Similarly, the architecture and the experimental setups follow VN-PointNet from \cite{deng2021vectorneurons}, and we replace a portion of vector channels with scalars in each layer. We evaluate the performance with different ratios of vector channels, where one vector channel is equivalent to three scalar channels. 
 
 In Table~\ref{tab:supp_hybrid}, we report the performance with different ratio of vector channels, where one vector channel is considered equivalent to three scalar channels. In both cases, our method with hybrid features achieves higher accuracy than pure vector features (100\%) as in \cite{deng2021vectorneurons}. The conclusion is consistent with the ablation experiments in Fig.~\ref{fig:ablation_channel} of the main paper, and our hybrid feature paradigm is advantageous in general cases.

\begin{table}[H]
\centering
\vspace{-1em}
\caption{\textbf{Ablation on vector vs. scalar channels.} We evaluate on ShapeNet surface reconstruction with non-graph structured equivariant implicit models, and ModelNet40 point cloud classification with equivariant point cloud networks, for which we follow the VN-PointNet and the VN-ONet architectures in \cite{deng2021vectorneurons} and use hybrid layers instead of pure vector neuron layers. 
}
\resizebox{0.89\columnwidth}{!}{%
\begin{tabular}{c| c c c c c c c}
\toprule
Ratio of vector channels & 0\% & 12.5\% & 25\% & 50\% & 75\% & 87.5\% & 100\% \\
\midrule
ShapeNet implicit surface reconstruction (mIoU) & 0.408 & 0.630 & 0.707 & {\bf 0.719} & {\bf 0.719} & 0.704 & 0.694 \\
ModelNet40 point cloud classification (mAcc) & 0.808 & 0.830 & 0.852 & {\bf 0.856} & 0.855 & 0.852 & 0.847 \\
\bottomrule
\end{tabular}
}
\label{tab:supp_hybrid}
\vspace{0.0em}
\end{table}

\subsection{Ablation on the architecture.} 
We ablate the implementation choices of our models. 
First, we explore how our models perform without the multi-scale sampling design on the ShapeNet object reconstruction and the Synthetic Room (SynRoom) scene reconstruction tasks. In Table~\ref{tab:supp_multiscale}, we show that the scene reconstruction performance drops more without the multi-scale architecture, while the difference in object reconstruction performance is subtle. We argue that scene reconstruction is a more complex task, so the multi-scale design plays a more important role to aggregate global and local context in different scales.

Then, we show the effect of using different point cloud encoders in our graph models and the baseline methods ConvONets~\cite{peng2020convolutional}. The results are in Table~\ref{tab:supp_encoder}. For the graph models, the scene-level reconstruction performance drops much more when the graph encoder is replaced by a PointNet encoder. The results indicate that both the locality modelling and the awareness of the translation equivariance from the graph encoder are more crucial for scene-level reconstruction as a more comlex task.
However, in ConvONets, using the graph point encoder instead of the PointNet encoder does not lead to a significantly improved performance. Unlike our graph methods, ConvONets learn the latent feature with an intermediate grid feature tensor. So
feature embedding in ConvONet relies more on the regular convolution layers applied on the grid feature, while the point encoder plays a less important role.
~
\begin{table}[t]
\centering
\caption{\textbf{Graph functions with and without multi-scale graph neighbor samplings.} The multi-scale structure improves more on scene reconstruction performance. 
}
\label{tab:supp_multiscale}
\resizebox{0.72\columnwidth}{!}{%
\begin{tabular}{c |c c| c c}
\toprule
Model    & \multicolumn{2}{c|}{GraphONet}  & \multicolumn{2}{c}{E-GraphONet} \\
Dataset    & ShapeNet & SynRoom & ShapeNet & SynRoom \\
\midrule
Multi-scale sampling & {\bf 0.904} &{\bf 0.883} & 0.890 & 0.848 \\
Single-scale sampling & 0.897 {\scriptsize [-0.007]} &0.859 {\scriptsize [-0.024]} & 0.884 {\scriptsize [-0.006]} & 0.814 {\scriptsize [-0.034]} \\
\bottomrule
\end{tabular}
}
\vspace{0.0em}
\end{table}
~
\begin{table}[t]
\centering
\caption{\textbf{Implicit functions with different point encoders.} Graph models with graph point encoders replaced by PointNets would lead to more performance drop on scenes than on objects, because PointNet models no locality or translation equivariance which are more crucial for scenes; ConvONets with different point encoders show similar performance, because in such methods, feature embedding relies more on the grid encoder than the point encoder. 
}
\label{tab:supp_encoder}
\resizebox{0.99\columnwidth}{!}{%
\begin{tabular}{c |c c| c c | c c | c c}
\toprule
Model    & \multicolumn{2}{c|}{GraphONet}  & \multicolumn{2}{c|}{E-GraphONet} & \multicolumn{2}{c|}{ConvONet-2D} & \multicolumn{2}{c}{ConvONet-3D} \\
Dataset    & ShapeNet & SynRoom & ShapeNet & SynRoom   & ShapeNet & SynRoom & ShapeNet & SynRoom \\
\midrule
Graph encoder &  {\bf 0.904} &{\bf 0.883} & 0.890 & 0.848 & 0.881 {\scriptsize [-0.003]} & 0.803  {\scriptsize [+0.001]}& 0.872  {\scriptsize [+0.002]}& 0.853  {\scriptsize [+0.006]}\\ 
PointNet encoder & 0.887 {\scriptsize [-0.017]} &0.826 {\scriptsize [-0.057]} & 0.879 {\scriptsize [-0.011]} & 0.797 {\scriptsize [-0.051]}  & 0.884 & 0.802 & 0.870 & 0.847 \\
\bottomrule
\end{tabular}
}
\vspace{0.0em}
\end{table}

\subsection{Learning curve with limited training data.} 
\begin{figure}[H]
\centering 
\vspace{1.2em}
\includegraphics[trim={0cm 0cm 0cm 0cm},clip,width=0.6\linewidth]{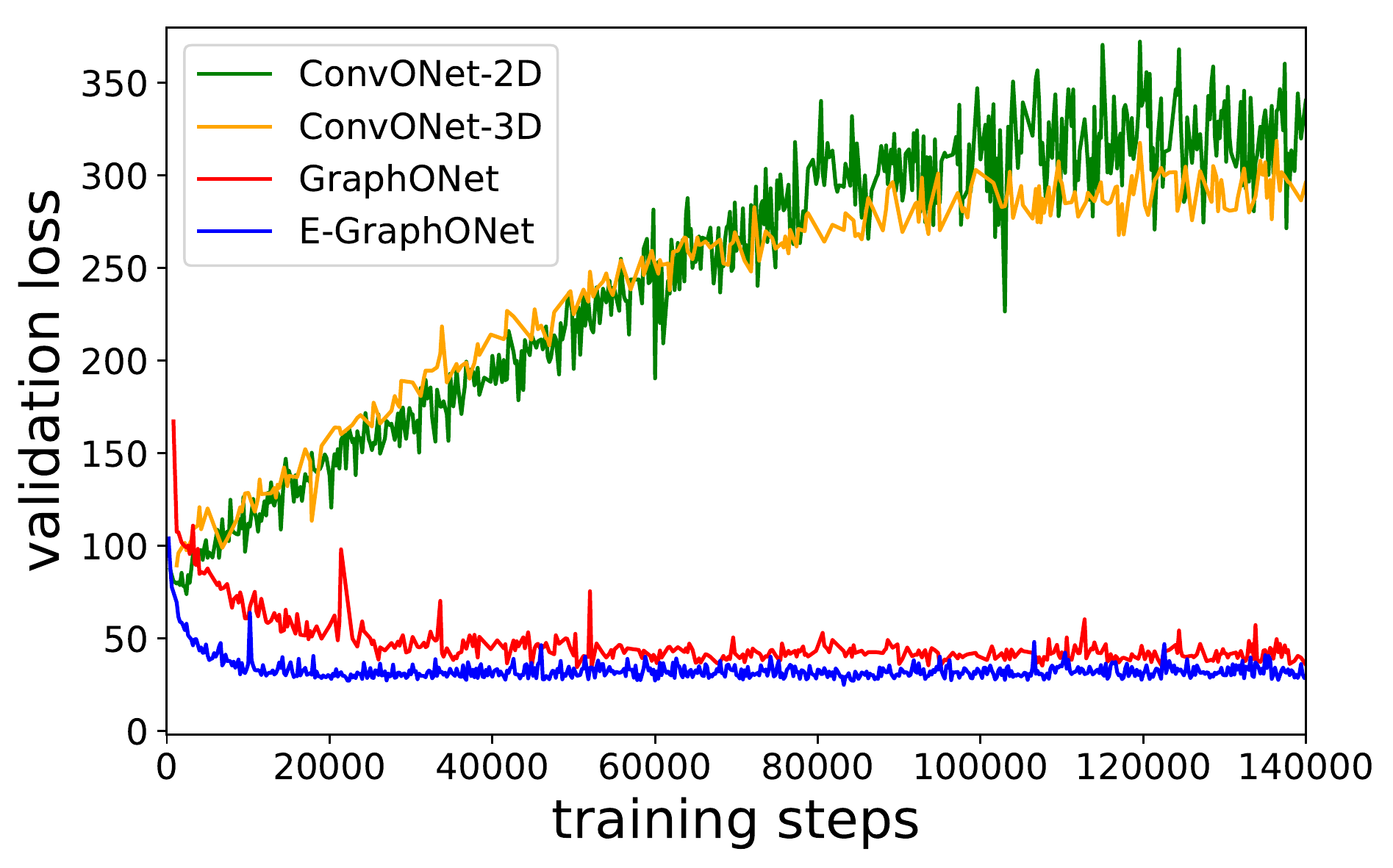}
\caption{\textbf{Learning curve with 130 training examples.} We show the validation loss curves. Graph methods can learn properly from very few training data, while ConvONets suffer from overfitting. 
}
\vspace{-1.0em}
\label{fig:supp_valloss}  
\end{figure}

We explore how the implicit model performs with very few training data of 130 examples, and provide the validation loss curve, as illustrated in Fig.~\ref{fig:supp_valloss}. This result is a supplement to the test performance in Table~\ref{tab:ablation_few} of the main paper.  Both ConvONet-2D and ConvONet-3D suffer from overfitting in the very early stage of training, prior to 500 training steps. By contrast, our graph methods are able to learn properly from very few training data. The equivariant model is with better validation loss and  more stable learning curve, which indicates that the equivariance property works as a regularization that controls the model complexity.

\subsection{More qualitative and quantitative results.}
 We evaluate ShapeNet reconstruction performance by each object category. The results are shown in Table~\ref{tab:supp_shapenet}, where our graph model shows better performance with most of the object categories. 
 
\begin{figure}[t]
\centering 
\vspace{-3.0em}
\includegraphics[trim={1.1cm 4.8cm 3.6cm 1cm},clip,width=1.0\linewidth]{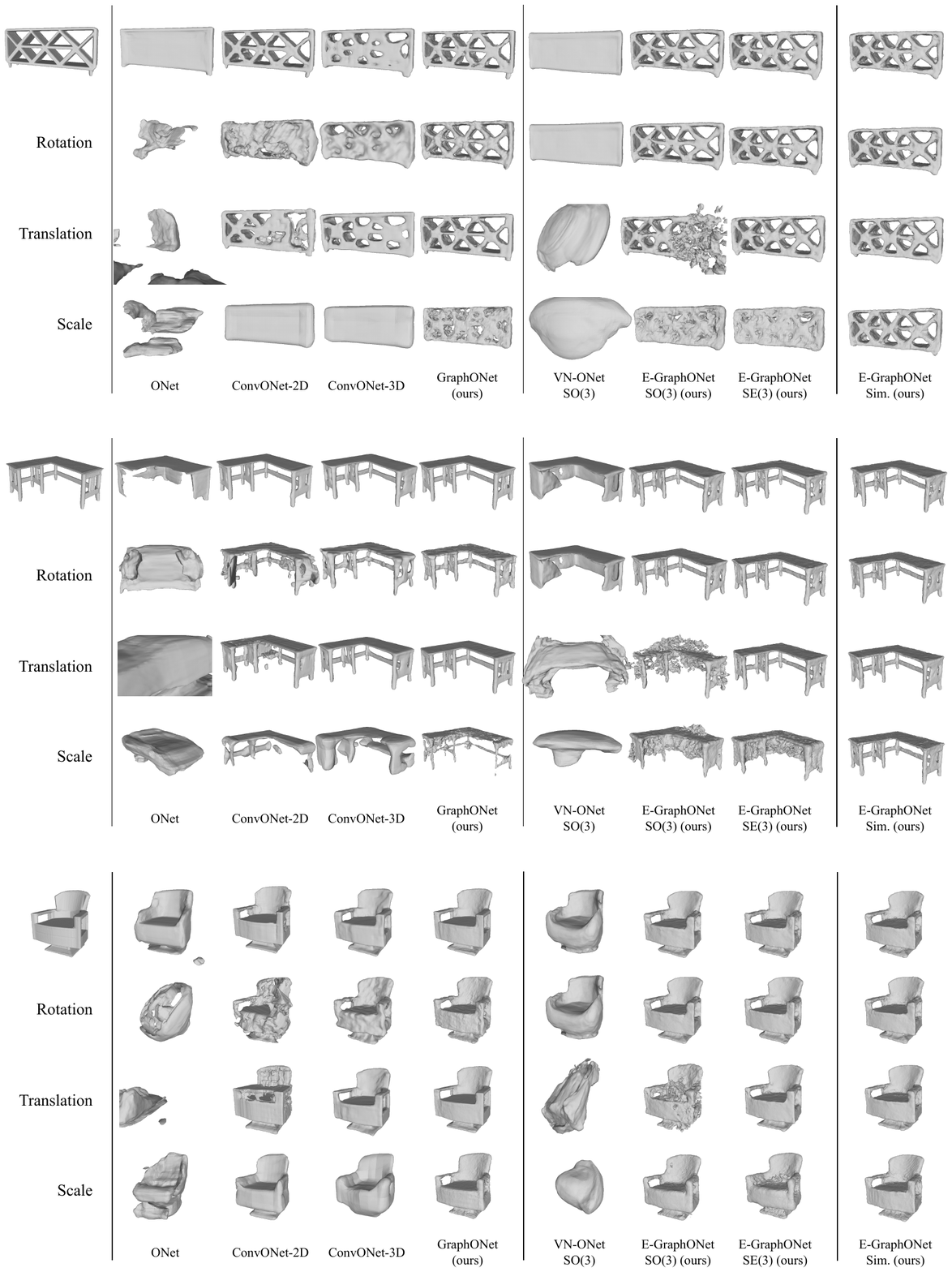}
\vspace{-0.6em}
\caption{\textbf{More examples on ShapeNet object reconstruction under unseen transformations.}
 With transformations we show the back-transformed shapes.}
\vspace{-0.1em}
\label{fig:supp_big}  
\end{figure}

 In Fig.~\ref{fig:supp_big}, we show some additional ShapeNet reconstruction examples under transformations. Our final equivariance model guarantees equivariance to all kinds of similarity transformations.   
 
\begin{figure}[t]
\centering 
\vspace{-1.0em}
\includegraphics[trim={1.3cm 3.4cm 1cm 1cm},clip,width=1.0\linewidth]{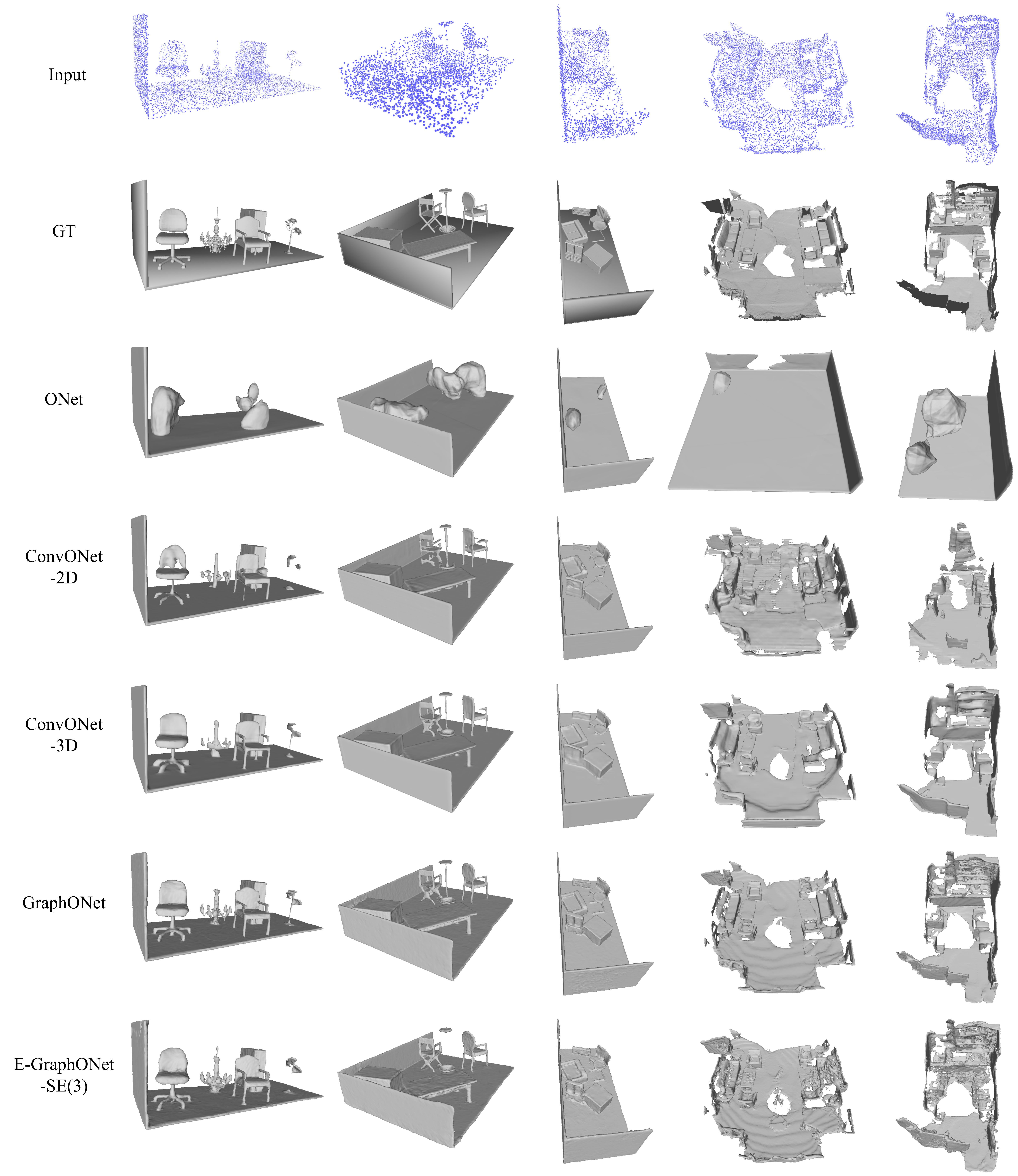}
\vspace{-0.6em}
\caption{\textbf{More scene reconstruction examples.} The left three columns are from the Synthetic Room dataset. The right two columns are from ScanNet.
}
\vspace{-0.1em}
\label{fig:supp_room}  
\end{figure}

 Though not the main focus of this paper, our method scale to scene-level reconstructions, and we show some more room reconstruction examples in Fig.~\ref{fig:supp_room}. 
Our GraphONet models better details. The equivariant model E-GraphONet achieves comparable numerical performance to that of ConvONet-3D, but the qualitative results are with some noisy artifacts, especially on the synthetic-to-real evaluation on the ScanNet dataset with corrupt areas in the point cloud scans. We argue that the restricted representation power of the equivariant layers limits the model to learn denoising and completion alongside reconstruction while generalize to more complex corrupted scenes. See the discussion on the limitation in the main paper.

Additionally, We include an appendix video at \url{https://staff.fnwi.uva.nl/y.chen3/3DEGIF/video.mp4} for some visualizations, especially for shape reconstructions under different poses and scales.

\begin{table}[H]
\vspace{-2.0em}
\caption{\textbf{Category-specific ShapeNet reconstruction performance.} We evaluate our methods and compare with the baseline implicit representation models. 
}
\centering
\vspace{0.0em}
\resizebox{1.02\columnwidth}{!}{%
\noindent
\begin{tabular}{c|ccc|ccc|ccc|ccc|ccc}
\toprule
            & \multicolumn{3}{c|}{ConvONet-2D~\cite{peng2020convolutional}}           & \multicolumn{3}{c|}{ConvONet-3D~\cite{peng2020convolutional}} & \multicolumn{3}{c|}{IF-Net~\cite{chibane2020ifnet}}       & \multicolumn{3}{c|}{GraphONet}                    & \multicolumn{3}{c}{E-GraphONet-SO(3)}    \\
            & IoU            & Chamfer & Normal         & IoU      & Chamfer   & Normal   & IoU   & Chamfer & Normal         & IoU            & Chamfer        & Normal         & IoU            & Chamfer        & Normal \\ \midrule
airplane    & 0.849          & 0.034   & 0.931          & 0.849    & 0.033     & 0.932    & 0.862 & 0.031   & 0.936          & \textbf{0.881} & \textbf{0.027} & \textbf{0.941} & 0.867          & 0.028          & 0.930  \\
bench       & 0.830          & 0.035   & 0.921          & 0.791    & 0.041     & 0.911    & 0.815 & 0.037   & 0.915          & \textbf{0.836} & \textbf{0.034} & \textbf{0.924} & 0.807          & 0.037          & 0.906  \\
cabinet     & 0.940          & 0.046   & 0.956          & 0.923    & 0.054     & 0.953    & 0.936 & 0.048   & 0.956          & \textbf{0.943} & \textbf{0.047} & \textbf{0.958} & 0.927          & 0.050          & 0.944  \\
car         & 0.886          & 0.075   & 0.893          & 0.877    & 0.080     & 0.891    & 0.890 & 0.072   & 0.894          & \textbf{0.897} & \textbf{0.068} & \textbf{0.895} & 0.890          & 0.072          & 0.886  \\
chair       & 0.871          & 0.046   & 0.943          & 0.853    & 0.049     & 0.942    & 0.878 & 0.043   & 0.946          & \textbf{0.895} & \textbf{0.039} & \textbf{0.951} & 0.879          & 0.043          & 0.939  \\
display     & 0.927          & 0.036   & 0.968          & 0.904    & 0.042     & 0.965    & 0.923 & 0.036   & 0.968          & \textbf{0.936} & \textbf{0.034} & \textbf{0.972} & 0.922          & 0.036          & 0.963  \\
lamp        & 0.785          & 0.059   & 0.900          & 0.792    & 0.066     & 0.910    & 0.820 & 0.047   & 0.916          & 0.847          & 0.042          & \textbf{0.922} & \textbf{0.848} & \textbf{0.040} & 0.915  \\
loudspeaker & 0.918          & 0.064   & 0.939          & 0.914    & 0.065     & 0.942    & 0.928 & 0.056   & 0.945          & \textbf{0.938} & \textbf{0.053} & \textbf{0.946} & 0.936          & 0.055          & 0.941  \\
rifle       & 0.846          & 0.028   & 0.929          & 0.826    & 0.031     & 0.924    & 0.842 & 0.028   & 0.928          & \textbf{0.877} & \textbf{0.022} & \textbf{0.943} & 0.868          & 0.023          & 0.933  \\
sofa        & 0.936          & 0.042   & 0.958          & 0.923    & 0.046     & 0.956    & 0.938 & 0.040   & 0.959          & \textbf{0.946} & \textbf{0.037} & \textbf{0.963} & 0.931          & 0.041          & 0.951  \\
table       & 0.888          & 0.038   & 0.959          & 0.860    & 0.043     & 0.956    & 0.880 & 0.038   & 0.959          & \textbf{0.896} & \textbf{0.036} & \textbf{0.963} & 0.869          & 0.040          & 0.950  \\
telephone   & \textbf{0.955} & 0.027   & \textbf{0.983} & 0.942    & 0.030     & 0.981    & 0.949 & 0.027   & \textbf{0.983} & 0.954          & \textbf{0.026} & \textbf{0.983} & 0.946          & 0.027          & 0.979  \\
vessel      & 0.865          & 0.043   & 0.919          & 0.860    & 0.045     & 0.919    & 0.876 & 0.040   & 0.923          & \textbf{0.901} & \textbf{0.033} & \textbf{0.934} & 0.892          & 0.035          & 0.924  \\ \midrule
mean        & 0.884          & 0.044   & 0.938          & 0.870    & 0.048     & 0.937    & 0.887 & 0.042   & 0.941          & \textbf{0.904} & \textbf{0.038} & \textbf{0.946} & 0.890          & 0.041          & 0.936 \\ \bottomrule
\end{tabular}
}
\label{tab:supp_shapenet}
\vspace{0.2em}
\end{table}

\end{document}